%% file: main.tex
\documentclass{bmvc2k}
\makeatletter
\let\BMVC@origmaketitle\maketitle
\renewcommand{\maketitle}{%
  \BMVC@origmaketitle%
  \let\maketitle\BMVC@origmaketitle
  \bmvaResetAuthors
}
\makeatother
\usepackage[dvipsnames]{xcolor}
\definecolor{cvprblue}{rgb}{0.21,0.49,0.74}
\hypersetup{allcolors=cvprblue}

\newcommand{\red}[1]{{\color{cvprblue}#1}}
\newcommand{\blue}[1]{{\color{cvprblue}#1}}
\newcommand{\todo}[1]{{\color{red}#1}}
\newcommand{\TODO}[1]{\textbf{\color{red}[TODO: #1]}}

\newcommand{\andreic}[1]{}
\newcommand{\andrei}[1]{#1}
\newcommand{\simonc}[1]{}
\newcommand{\simon}[1]{#1}

\usepackage{cleveref}
\usepackage{comment}
\usepackage{booktabs}
\usepackage[table]{xcolor}
\usepackage{multirow} 
\usepackage{wrapfig}

\usepackage{pifont}% http://ctan.org/pkg/pifont
\newcommand{\checkmark}{\ding{51}}%
\newcommand{\xmark}{\ding{55}}%
\usepackage{amssymb}

\title{LED: Light Enhanced Depth Estimation at Night}

\addauthor{Simon de Moreau}{https://simon.demoreau.fr/}{1,2}
\addauthor{Yasser Almehio}{yasser.almehio@valeo.com}{2}
\addauthor{Andrei Bursuc}{https://abursuc.github.io/}{3}
\addauthor{Hafid El-Idrissi}{hafid.el-idrissi@valeo.com}{2}
\addauthor{Bogdan Stanciulescu}{bogdan.stanciulescu@minesparis.psl.eu}{1}
\addauthor{Fabien Moutarde}{https://people.minesparis.psl.eu/fabien.moutarde/}{1}

\addinstitution{
 Mines Paris - PSL University\\
 Paris, France
}
\addinstitution{
 Valeo\\
 Paris, France
}
\addinstitution{
 Valeo AI\\
 Paris, France
}

\runninghead{de Moreau \etal}{LED: Light Enhanced Depth Estimation at Night}

\def\eg{\emph{e.g}\bmvaOneDot}
\def\Eg{\emph{E.g}\bmvaOneDot}
\def\etal{\emph{et al}\bmvaOneDot}

\begin{document}

\maketitle

\begin{abstract}
Nighttime camera-based depth estimation is a highly challenging task, especially for autonomous driving applications, where accurate depth perception is essential for ensuring safe navigation.
\simon{Models trained on daytime data often fail in the absence of precise but costly LiDAR. Even vision foundation models trained on large amounts of data are unreliable in low-light conditions. 
In this work, we aim to improve the reliability of perception systems at night time.
To this end, }we introduce Light Enhanced Depth (LED), a novel, cost-effective approach that significantly improves depth estimation in low-light environments by harnessing a pattern projected by high definition headlights available in modern vehicles. 
LED leads to significant performance boosts across multiple depth-estimation architectures (encoder-decoder, Adabins, DepthFormer, Depth Anything V2) both on synthetic and real datasets.
Furthermore, increased performances beyond illuminated areas reveal a holistic enhancement in scene understanding.
Finally, we release the Nighttime Synthetic Drive Dataset, a synthetic and photo-realistic nighttime dataset, which comprises 49,990 comprehensively annotated images. 
%\vspace{30pt}
To facilitate further research, both synthetic dataset and code are publicly available at \href{https://simondemoreau.github.io/LED/}{https://simondemoreau.github.io/LED/}.
\end{abstract}

%-------------------------------------------------------------------------

\input{sections/intro}

\input{sections/rw}

\input{sections/method}

\input{sections/dataset}

\input{sections/experiments}
\input{sections/limitations_conclusion}

\bibliography{main}
\clearpage

\input{supplementary}

\end{document}

%% file: sections/intro.tex
%\vspace{-8pt}
\section{Introduction}
%\vspace{-3pt}
\label{sec:intro}
\begin{figure}[t]
    \centering
    \includegraphics[width=0.7\textwidth]{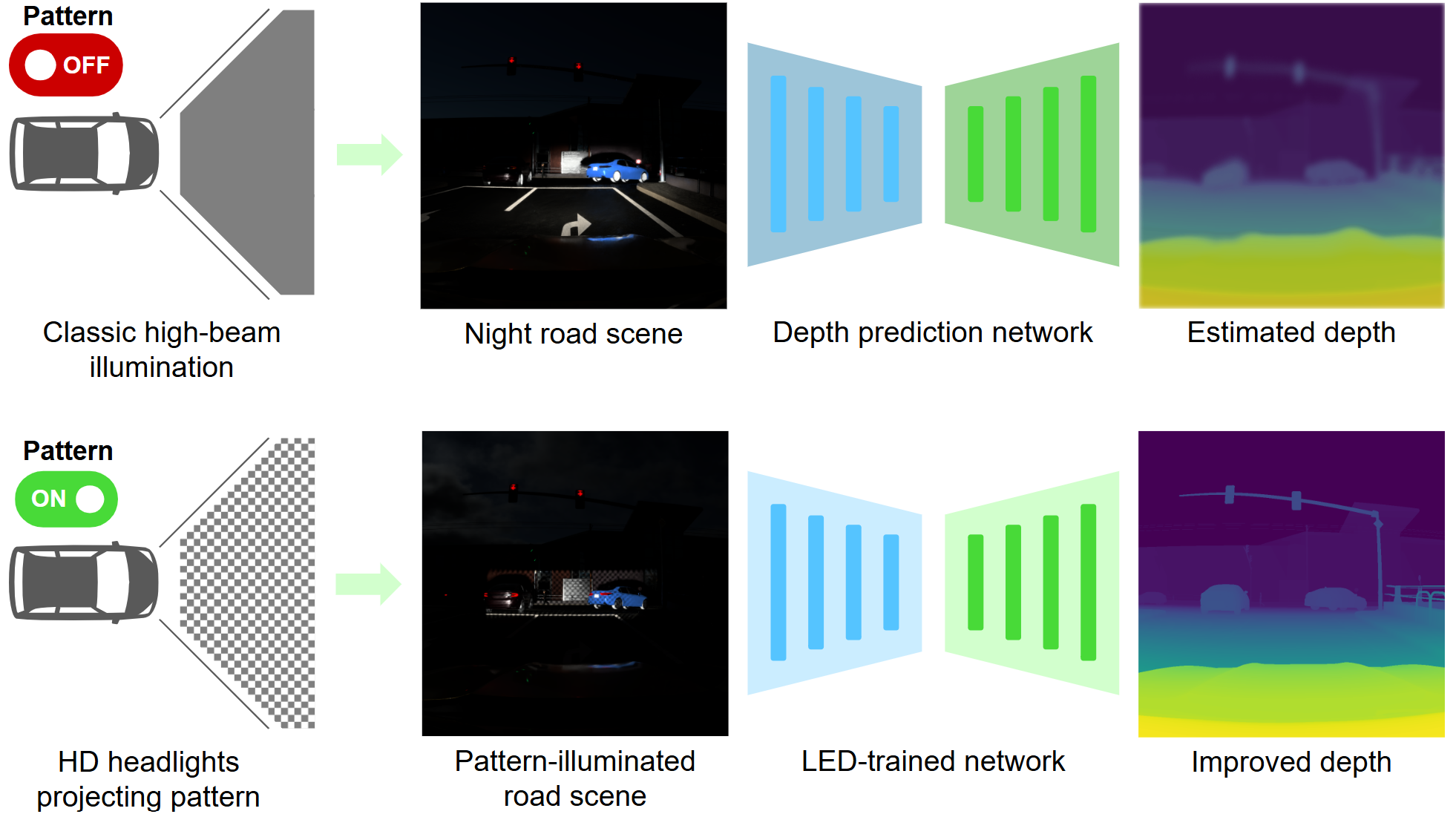}
    \caption{\simon{\textbf{Light Enhanced Depth (LED)} is a novel method that harnesses high-definition headlights' pattern projected onto the scene to enhance nighttime depth estimation from RGB images.
    We release a synthetic nighttime dataset with high beam and pattern-illuminated images, along with comprehensive ground truth annotations, to advance research in nighttime perception.
    }
    }
    \label{fig:concept-fig}
    \vspace{-15pt}
\end{figure}

Adverse conditions, such as harsh weather or nighttime, pose significant challenges to many computer vision applications. Despite impressive progress in perception systems for autonomous driving, enabled by powerful deep neural architectures and training techniques, the challenges of nighttime navigation persist. 
Accurate depth estimation is a crucial aspect of perception, profoundly impacting overall scene perception and comprehension, especially at night~\cite{silberman_indoor_2011}. While LiDAR sensors offer high accuracy, their widespread adoption is impeded by their substantial cost. Additionally, in humid weather, LiDAR effectiveness can be significantly degraded due to beam reflection and increased noise levels~\cite{bijelic2018benchmark, bijelic2018benchmarking}.

Cameras are powerful, cost-effective sensors that capture rich information about the environment and are deployed in most modern vehicles. Camera-based perception solutions display high reliability and accuracy 
\cite{godard2019digging, bian2019unsupervised, guizilini20203d, godard2017unsupervised, bhat2021adabins, agarwal2023attention, li2023depthformer}, but mostly on clear daylight conditions: they struggle under distribution shift and low-light conditions. While foundation models \cite{depth_anything_v1,depth_anything_v2,depth_pro,marigold} show great improvement for depth estimation on multiple domains, they are still not robust to nighttime images 
that are long-tail in the training data distribution.
\simon{The availability of labeled data necessary for training high-capacity deep neural networks is an additional challenge.}
For the task of depth estimation, some methods rely on supervised learning \cite{bhat2021adabins, agarwal2023attention, li2023depthformer, lee2019big}, but most use self-supervised methods \cite{godard2019digging, poggi2018towards, zhou2017unsupervised, yin2018geonet, shu2020feature, bian2019unsupervised, guizilini20203d, wang2021regularizing, zheng2023steps} since obtaining ground truth depth information is expensive.
While dedicated approaches for specific conditions have emerged in recent years \cite{wang2021regularizing, zheng2023steps, vankadari2020unsupervised, spencer2020defeat, liu2021self}, they still only partially mitigate these issues.
To the best of our knowledge, they remain limited to self-supervised strategies, due to the lack of publicly available, large-scale annotated datasets suitable for nighttime depth estimation. 
\simon{To address this gap, this paper releases a synthetic nighttime dataset annotated with dense depth maps, along with additional labels.}

\simon{
In this paper, we introduce Light Enhanced Depth (LED), a novel approach that significantly improves depth estimation in low-light environments, ensuring enhanced accuracy and reliability for autonomous vehicles. 
High-Definition (HD) headlights, commonly found in modern vehicles, have shown promising results in scene perception research \cite{waldner_hardware---loop-simulation_2019, waldner_digitization_2020, waldner_optimal_2021}. Drawing inspiration from active stereovision \cite{fanello_hyperdepth_2016,riegler_connecting_2019, li_deep_2022,baek_polka_2021}, we harness HD headlights to project a pattern into the scene, guiding the network and thereby improving performance.
}

Our contributions can be summarized as follows:\\
{\small$\bullet$} \textbf{Architecture-agnostic enhancement}: 
LED can be applied to any depth estimation architecture to improve nighttime performances (RMSE: -11\% on encoder-decoder, -24.06\% on Adabins \cite{bhat2021adabins}, -8.00\% on DepthFormer \cite{li2023depthformer}, -15.5\% on Depth Anything V2 \cite{depth_anything_v2}). \\
{\small$\bullet$}  \textbf{Data-efficiency}: 
LED-trained models outperform the others with only 20\% of training data. \\
{\small$\bullet$}  \textbf{Real prototype}: LED demonstrates great performances improvements on our in-house dataset, collected using a real car-mounted prototype. \\
{\small$\bullet$}  \textbf{Dataset}: We provide the Nighttime Synthetic Drive Dataset, a photorealistic synthetic dataset, comprising 49,995 comprehensively annotated nighttime images to foster future investigations in nighttime perception. 
%\newpage

%% file: sections/rw.tex
\vspace{-6pt}
\section{Related Work}
%\vspace{-8pt}
\subsection{Depth Estimation from Camera}
%\vspace{-3pt}
\textbf{Supervised Learning.} Recent advances in supervised depth prediction leverage %exploit
transformer architectures and attention mechanisms \cite{bhat2021adabins, agarwal2023attention, li2023depthformer, lee2019big} to improve performances. 
The use of bins \cite{bhat2021adabins,agarwal2023attention} improves depth prediction accuracy even with limited datasets available \cite{Geiger2012CVPR, nuscenes, RobotCarDatasetIJRR, Silberman:ECCV12} compared to self-supervised methods \cite{godard2019digging,poggi2018towards,zhou2017unsupervised,yin2018geonet,bian2019unsupervised,guizilini20203d,shu2020feature}. However, these methods are confined to daytime scenarios. 
Nighttime depth estimation approaches \cite{wang2021regularizing, zheng2023steps} \simon{rely on self-supervised methods, due to the lack of large, annotated nighttime datasets.}
In response, we release the Nighttime Synthetic Drive Dataset to support \simon{the development of nighttime-specific methods.} \\
\textbf{Vision Foundation Models.} 
\simon{Depth foundation models \cite{depth_anything_v1,depth_anything_v2,depth_pro,marigold} have achieved impressive generalization by leveraging extensive synthetic and real-world training data. However, we show in \cref{sec:depth-any} that these models perform poorly in zero-shot evaluations on our nighttime dataset. It highlights the need for models specialized in nighttime conditions and underscores the benefits of our method.} \\
\simon{\textbf{Nighttime Domain Adaptation.}
Nighttime depth estimation can be approached as a domain adaptation problem \cite{gasperini_morbitzer2023md4all}, 
focusing on aligning features between daytime and nighttime images
\cite{vankadari2020unsupervised, spencer2020defeat,liu2021self,liu2021self}.\,\,Our method specifically adapts to nighttime scenes features by leveraging the informative deformations of the light pattern emitted by a vehicle’s HD headlamp.}

%\vspace{-12pt}
\subsection{Light for Perception}
%\vspace{-3pt}
\textbf{Active Stereovision.}
Depth estimation based on active stereovision involves disparity measurement. Unlike traditional stereovision, disparity is computed between an image and a pattern projected onto the scene. Recent methods \cite{fanello_hyperdepth_2016,riegler_connecting_2019, li_deep_2022,baek_polka_2021} use deep learning models designed to take patterns and images as input. 
While active stereovision offers valuable scale information, it has demonstrated significant performance degradation when deployed outdoors, due to high ambient lighting and the low power of projectors \cite{mertz_low-power_2012, gupta_structured_2013, brahmanage_outdoor_2019}. \simon{We avoid this thanks to our nighttime environment and the utilization of high-powered HD headlights.}\\
\textbf{HD Lighting.}
A series of works \cite{de_charette_fast_2012, fleet_programmable_2014} highlight \simon{HD lighting} potential applications and particularly its ability to design anti-glare systems while optimizing illumination for enhanced driver visibility\simon{, even in adverse conditions such as rain or snow.}
More recently, \cite{waldner_energy-efficient_2022} propose reducing overall scene illumination to decrease power consumption while maintaining object detection capabilities. 
\simon{\cite{waldner_hardware---loop-simulation_2019, waldner_digitization_2020, waldner_optimal_2021} also} have proposed hardware-in-the-loop simulation for HD headlights. 
We introduce a novel application of HD headlights to improve understanding of overall scene geometry and enhance depth estimation. 
We achieve this by projecting a pattern onto the nighttime scene, providing guidance for the model.

%% file: sections/method.tex
%\vspace*{-8pt}
\section{Method}
%\vspace{-5pt}
We improve depth estimation from monocular camera at nighttime by leveraging a pattern projected in front of the vehicle with HD headlights.

\vspace{-6pt}
\subsection{HD Pattern and Headlights}
%\vspace{-5pt}

Networks designed for active stereovision exploit the disparity between the projected pattern and the camera's view to estimate distances within the scene. With LED, a model identifies pattern's areas that deviate from the implicitly learned reference pattern.

\begin{wrapfigure}{l}{0.5\textwidth}
    \vspace{-0pt}
    \centering
    \includegraphics[width=0.343\textwidth]{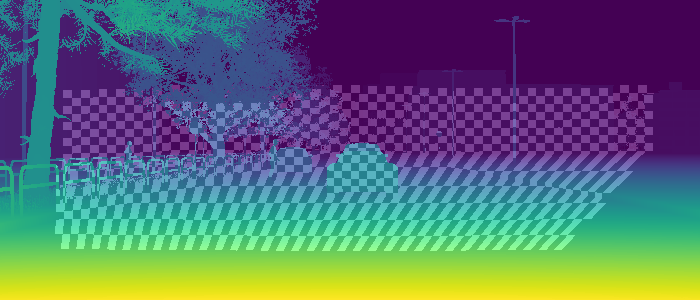}
    \includegraphics[width=0.147\textwidth]{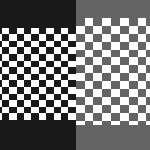}

    \caption{\textbf{Pattern deformation in the scene:} \simon{Left: For explanatory purpose, we project the pattern on a depth image.}
    It demonstrates trapezoidal deformations on horizontal surfaces and undistorted squares on vertical surfaces. Complex deformations on the car reveal insights into its geometry. \simon{Right:} Pattern projected on a wall at 10\,m \simon{(darker)} and 100\,m \simon{(lighter)}. Square sizes increases with distance.}
    \label{fig:hd-proj}
    \vspace{-5pt}
    
\end{wrapfigure}
%\vspace{-5pt}
\simon{Unlike traditional active stereovision, which relies on infrared lasers to project points, we harness an HD headlight.} 
Similar to a projector, this headlight can dynamically project any image or shape onto the scene.
\simon{To prevent pattern overlap, we use only the left headlight.

We employ a regular checkerboard pattern, because of its high contrast and sharp discontinuities. The dense concentration of corners and transitions makes it easily detectable by convolutional neural networks \cite{baker2020local}. 
Upon projection, the checkerboard is deformed according to surface shapes (from the camera's perspective). Horizontal planes stretch the pattern into a trapezoid, while vertical surfaces parallel to the image plane cause no deformation (see \cref{fig:hd-proj}). Non-planar surfaces, such as cars, result in more complex distortions. These deformations provide crucial geometric cues to the network, improving depth estimation accuracy.} 
Additionally, the emitted light is not colimated: \simon{as the projection extends further, the light becomes more divergent, causing pixels to appear larger (see \cref{fig:hd-proj}), thus providing useful depth cues to the network. } 

\vspace{-6pt}
\subsection{Architectures}
\label{sec:archi}
%\vspace{-3pt}
\simon{Our method is architecture-agnostic. By relying on a single pattern, the model can implicitly learn it, eliminating the need for a dedicated network architecture.}
To demonstrate the benefits of projecting an HD pattern for depth estimation, we leverage a simple encoder-decoder model~\cite{U-Net}. 
In spite of its straight-forward architecture, it achieves results on par with or superior to other state-of-the-art (SOTA) methods (see \cref{sec:hd_pattern_impact}). We also successfully apply our method to more complex SOTA architectures, 
such as DepthFormer~\cite{li2023depthformer} and Adabins~\cite{bhat2021adabins}, with similar performance gains (see \cref{sec:hd_pattern_impact}). 
\simon{Finally, we show that finetuning Depth Anything V2 \cite{depth_anything_v2} with LED, on a reduced amount of training samples, enable robust and precise depth estimation.}

%\vspace{-13pt}
\subsection{Implementation Details} 
%\vspace{-5pt}
\simon{
We implement the encoder-decoder in PyTorch \cite{Pytorch_NEURIPS2019_9015}. 
The model is trained from scratch during 70 epochs using the AdamW optimizer \cite{loshchilov2017decoupled}, with a batch size of 32 and a learning rate of $10^{-3}$.
Input and output resolutions are set to 320$\times$320\,px. 
Similar to \cite{HigherResMap}, our learning objective combines losses based on Log L1, gradient and normals. More information is available in the supplementary. 
The selection of the best epoch is based on the Root Mean Square Error (RMSE).
The training process takes 4 hours on a single Nvidia RTX 4090. }\

\simon{We implement Adabins \cite{bhat2021adabins} using the official code and rely on the toolbox \cite{lidepthtoolbox2022} for DepthFormer \cite{li2023depthformer}. We apply Depth Anything V2 \cite{depth_anything_v2} official fine-tuning code to adapt the model for metric depth estimation. 
The only changes are limited to dataset format compatibility. The training setup follows the original papers’ recommendations, excluding data augmentation to preserve the HD pattern from cropping.
}

\begin{figure*}[t]
    \centering
    \includegraphics[width=\textwidth]{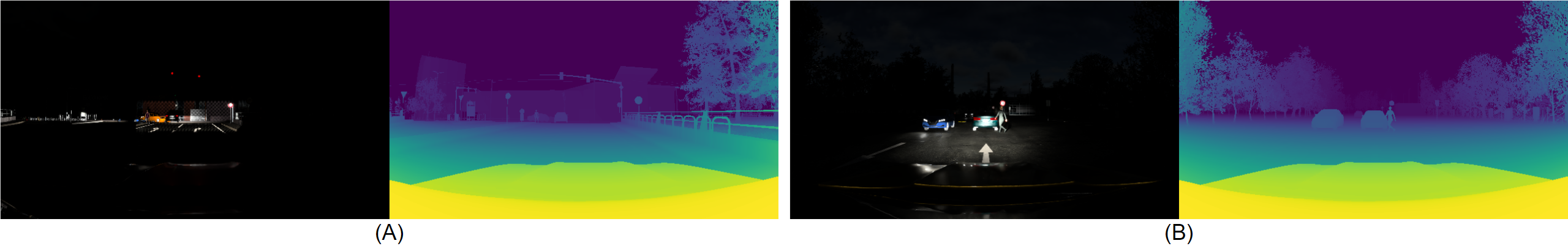}
    %\vspace*{-20pt}
   \caption{\textbf{Nighttime Synthetic Drive Dataset examples:} (A) depicts HD pattern and (B) high-beam illumination. Ground truth annotations include dense depth maps, semantic segmentation, instance segmentation labels and bounding boxes.}
   
    \label{fig:dataset_show}

    \vspace{-15pt}

\end{figure*}

%% file: sections/dataset.tex
\vspace{-6pt}
\section{Dataset}
%\vspace{-8pt}
\subsection{Nighttime Synthetic Drive Dataset (NSDD)} 
\label{sec:NSDD}

    \simon{Due to the lack of public datasets containing HD pattern illumination, we create the Nighttime Synthetic Drive Dataset (NSDD) using the Nvidia Drive Sim (Drop 15) simulator \cite{drivesim}, which generates road images with photorealism effort. To simulate realistic headlight projections, we adapt the vehicle's headlights based on photometric measurements. The dataset includes 24,995 images with pattern and 24,995 images with high beam illumination (see \cref{fig:dataset_show}).
    We simulate a real HD headlight with a resolution of 132$\times$28\,px and a field of view of  35°$\times$7°. The checkerboard cells are 0.5°, ensuring visibility with the camera resolution.} 
    
    \simon{We choose High Beam (HB) as the comparison baseline, considering it the maximum normal lighting condition. Since high beam illumination exceeds that of the checkerboard pattern, performance improvements are attributed solely to the pattern’s contribution.}
    \simon{In each frame, vehicles, pedestrians, and traffic signs are randomized. To simulate light interference, we include other car headlights and randomize ambient light levels between 0 and 10 lux.}
    \begin{wrapfigure}{l}{0.5\textwidth}
    \vspace{4pt}
    \centering
    \includegraphics[width=0.24\columnwidth]{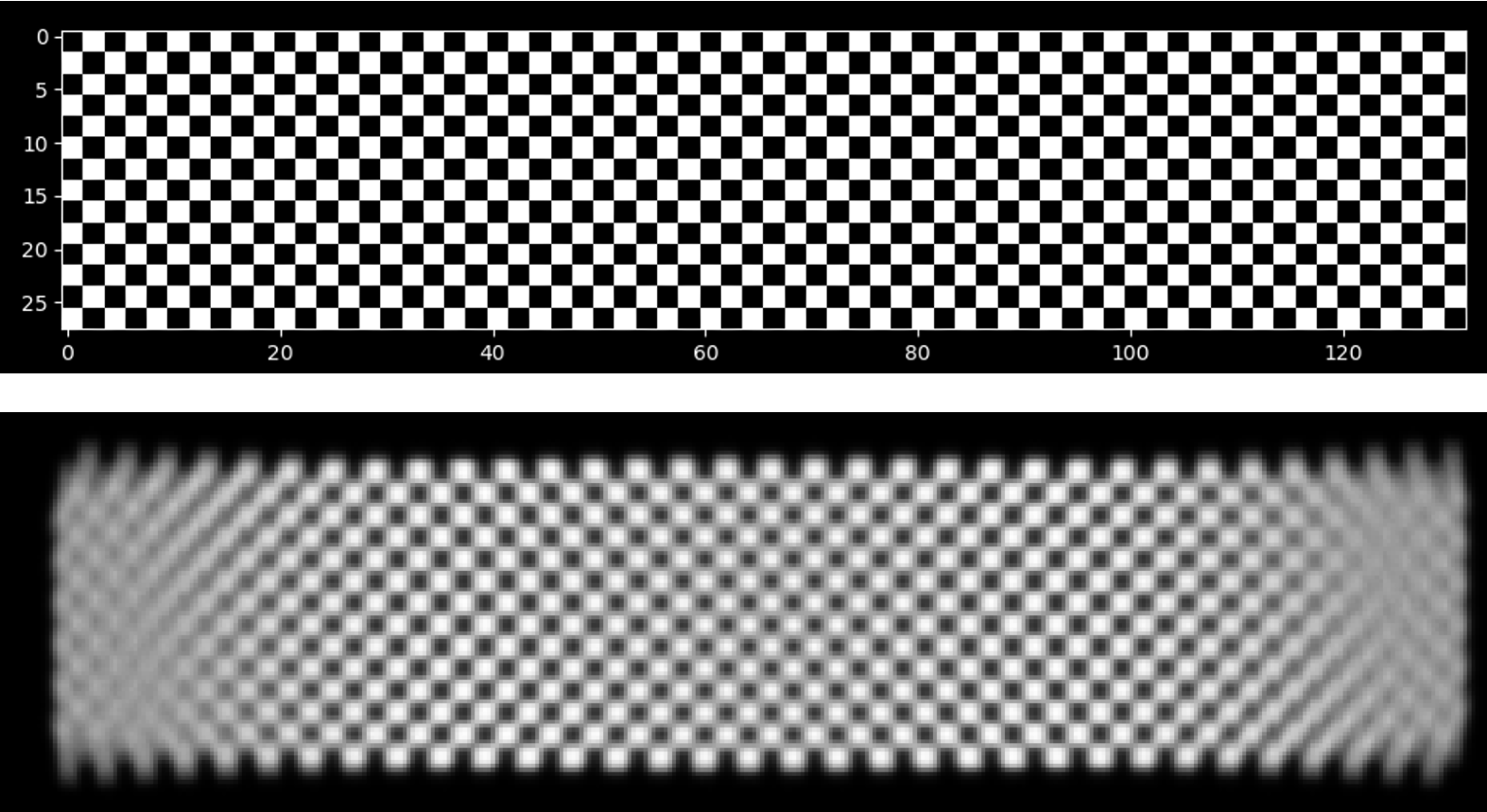}
    \includegraphics[width=0.24\columnwidth]{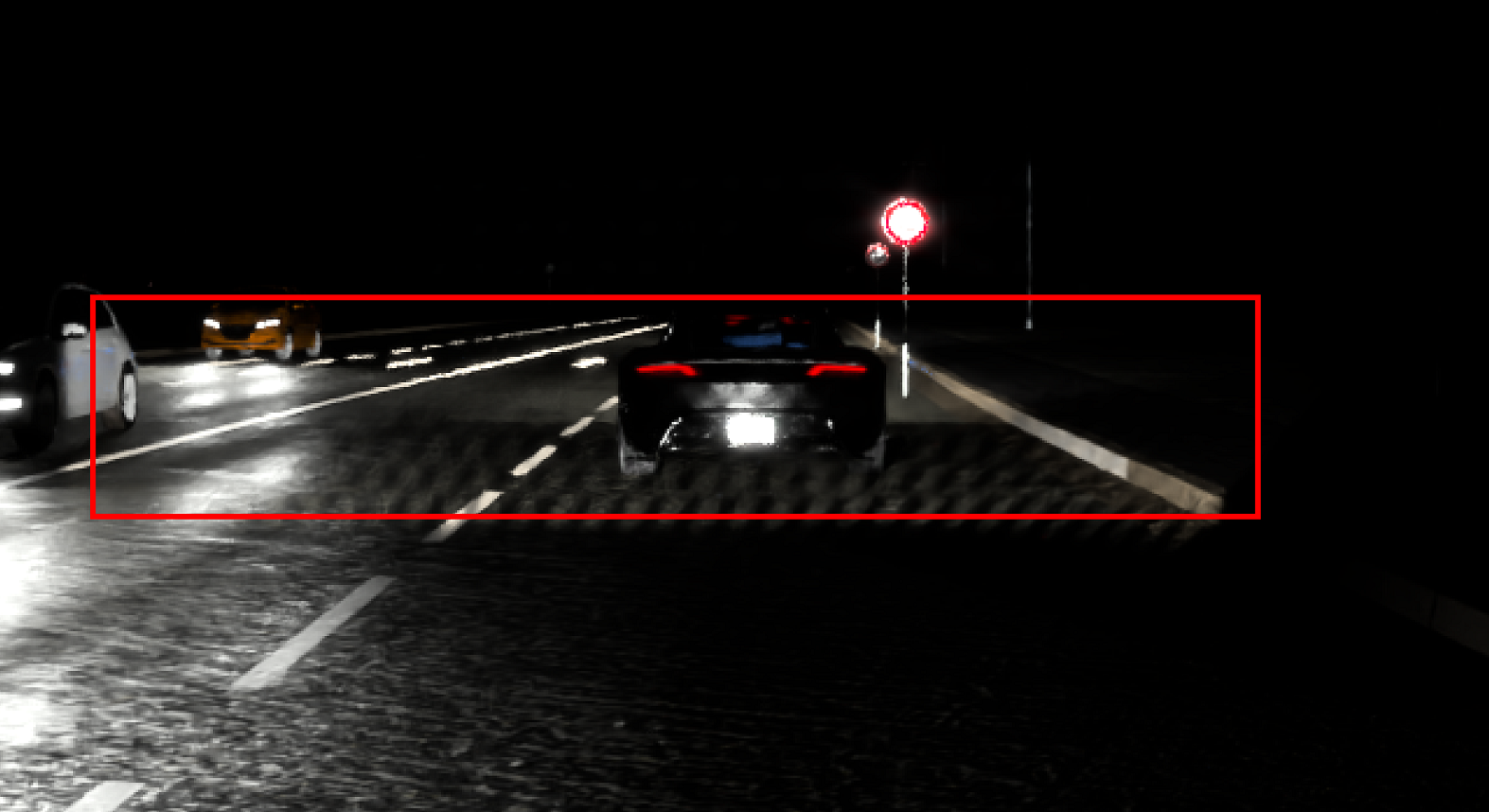}
    \caption{\textbf{Simulation of HD pattern:} Top-left: control matrix of the HD headlight. % (black=0\%, white=100\% intensity). 
    Bottom-left: photometry considering aberrations created by the headlight lens. Right: Resulting image using the photometry. 
    The area outlined in red \simon{is} the region of interest.}
    \label{fig:pattern_simu}
    \vspace{-4pt}
    \end{wrapfigure} 
    
    \simon{The dataset consists of 5 different maps (3 train / 1 val / 1 test), with 4,999 frames generated for both pattern and HB illuminated images in each. These maps represent major cities. Both parts of the dataset share the same randomization code, making the domains comparable.}

    \simon{To closely replicate reality, we account for the optical imperfections of headlights using real measured photometry data (see \cref{fig:pattern_simu}).}
    \simon{Aberrations observed in the pattern, particularly along the edges of the projection, are due to these imperfections.}
    \simon{All images have a resolution of 1920$\times$1080\,px, and we provide depth maps, annotations for 2D/3D object detection, normal estimation, and both semantic and instance segmentation.}
    \simon{The provided depth values are not limited, though they are clipped at 100\,m for all experiments. }
    During training, a 640$\times$640\,px square is initially center-cropped and then resized to 320$\times$320\,px.
    To ensure reproducibility, code and synthetic dataset are publicly available. Releasing large-scale nighttime data, with precise and comprehensive annotations, will foster research in nighttime computer vision for autonomous driving. 

\vspace{-6pt}
\subsection{Real-world Dataset}
%\vspace{-3pt}
    \simon{We validate LED on a real-world, in-house dataset collected using a car-mounted prototype.}
    It comprises 50,000 images (70\% train / 15\% val / 15\% test) from populated urban and rural roads, evenly split between Low Beam (LB) illumination and checkerboard pattern with 0.5°, 0.25° and 0.125° cells' dimension. 
    \simon{Ground truth is obtained from LiDAR data using the DOC-Depth method \cite{de2025doc} and Exwayz software \cite{exwayz_3dm}.}
    More details are available in supplementary materials.

%% file: sections/experiments.tex
\vspace{-6pt}
\section{Experiments}
%\vspace{-3pt}

\begin{figure*}[t]
    \centering
    \includegraphics[width=0.65\linewidth]{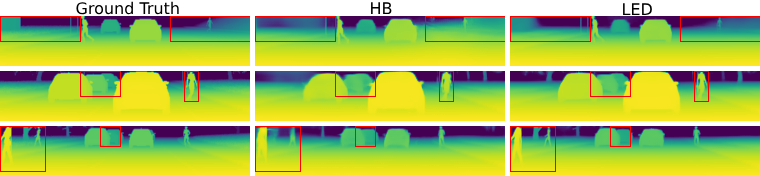}
    \caption{\textbf{Qualitative results on NSDD.} LED results %(right) 
    exhibit higher precision, \simon{and} more accurate object boundaries and shapes compared to HB. % (middle). 
    Red boxes indicate enhanced regions.} 
    \label{fig:qualitatif-results}
    %\vspace{-10pt}
\end{figure*}

\begin{table*}[tb]
\setlength{\tabcolsep}{8pt} % Default value: 6pt
\renewcommand{\arraystretch}{1.25} % Default value: 1
    \centering

\resizebox{0.65\textwidth}{!}{
    \begin{tabular}{l  c c c c c c c c c} \toprule 
    
         Pattern &  RMSE $\downarrow$&  Abs Rel $\downarrow$&  $\text{Log}_{10}$ $\downarrow$&  RMSE Log $\downarrow$&  SILog $\downarrow$&  Sq Rel $\downarrow$&  $\delta^1$ $\uparrow$&  $\delta^2$ $\uparrow$& $\delta^3$ $\uparrow$\\ 
         % \hline \hline 
         \midrule         
         \multicolumn{10}{l}{\cellcolor{gray!20}\emph{Encoder-decoder}} \\
         HB&  6.1204&  \textbf{0.0903}&  0.0233&  0.1353&  13.4847&  \textbf{3.3579}&  0.9489&  0.9812& 0.9908\\ 
         LED&  \textbf{5.4259}&  0.1996&  \textbf{0.0188}&  \textbf{0.1253}&  \textbf{12.4900}&  16.1224&  \textbf{0.9603}&  \textbf{0.9846}& \textbf{0.9927}\\ 
         % \hline
         \midrule
         \multicolumn{10}{l}{\cellcolor{gray!20}\emph{Adabins}} \\
         HB&  7.3520&  0.1360&  0.0320&  0.1230&  10.1880&  4.2770&  0.8790&  0.9440& 0.9740\\ 
         LED&  \textbf{5.5830}&  \textbf{0.0690}&  \textbf{0.0240}&  \textbf{0.0920}&  \textbf{7.7290}&  \textbf{2.5090}&  \textbf{0.9210}&  \textbf{0.9680}& \textbf{0.9840}\\ 
         % \hline 
         \midrule
         \multicolumn{10}{l}{\cellcolor{gray!20}\emph{DepthFormer}} \\
         HB&  5.8528&  \textbf{0.0687}&  0.0281&  0.1205&  11.3841&  \textbf{0.8328}&  0.9432&  0.9802& 0.9913\\ 
         LED& \textbf{5.3845}& 0.1215& \textbf{0.0276}& \textbf{0.1111}& \textbf{10.4146}& 3.4015&\textbf{ 0.9497}& \textbf{0.9838}&\textbf{0.9950}\\ 
         % \hline
         \bottomrule
    
    \end{tabular}
    }
    
    \caption{\textbf{Comparison of depth estimation performances on NSDD:} HB models are trained on high beam data and LED on HD light pattern. LED models outperform HB models across all metrics, exceptions are for Abs Rel and Sq Rel. Metrics are computed in the ROI.}
\label{tab:res_HP}
\vspace{-15pt}
\end{table*}

We evaluate our method through extensive experiments on the Nighttime Synthetic Drive Dataset. 
We validate the contribution of our light pattern in boosting performance, both inside and outside illuminated area. 
We apply our method to the encoder-decoder and other SOTA approaches: Adabins \cite{bhat2021adabins}, DepthFormer \cite{li2023depthformer} and Depth Anything V2 \cite{depth_anything_v2}, showing its agnosticity.
Finally, we evaluate LED robustness beyond its training domain and its capabilities on real-world scenarios. 
\simon{Metrics used align with prior works \cite{bhat2021adabins,li2023depthformer,spencer2020defeat,liu2021self,wang2021regularizing,zheng2023steps}.}

\begin{table*}[t]
\setlength{\tabcolsep}{8pt} % Default value: 6pt
\renewcommand{\arraystretch}{1.25} % Default value: 1
    \centering

\resizebox{0.65\textwidth}{!}{
    \begin{tabular}{c | c c c c c c c c c c } 
    % \hline 
    \toprule

          Model &  RMSE $\downarrow$&  Abs Rel $\downarrow$&  $\text{Log}_{10}$ $\downarrow$&  RMSE Log $\downarrow$&  SILog $\downarrow$&  Sq Rel $\downarrow$&  $\delta^1$ $\uparrow$&  $\delta^2$ $\uparrow$& $\delta^3$ $\uparrow$\\ 
          % \hline \hline 
          \midrule
         (HB) O-ROI&  9.2533&  0.1598&  0.0249&  0.2047&  20.4447&  11.4068&  0.9385&  0.9714& \textbf{0.9828}\\ 
         (LED) O-ROI&  \textbf{9.0070}&  \textbf{0.1295}&  \textbf{0.0221}&  \textbf{0.2027}&  \textbf{20.2348}&  \textbf{8.7407}&  \textbf{0.9427}&  \textbf{0.9723}& \textbf{0.9828}\\ 
         % \hline
         \midrule
         (HB) Full&  8.9702&  0.1521&  0.0247&  0.1988&  19.8572&  10.5225&  0.9396&  0.9725& 0.9836\\ 
         (LED) Full&  \textbf{8.6963}&  \textbf{0.1371}&  \textbf{0.0217}&  \textbf{0.1965}&  \textbf{19.6137}&  \textbf{9.5517}&  \textbf{0.9446}&  \textbf{0.9737}& \textbf{0.9839}\\
         % \hline
         \bottomrule

    \end{tabular}
    }
    \caption{\textbf{Encoder-decoder performance beyond ROI on NSDD:} O-ROI stands for Outside ROI, where the evaluation mask is the inverse of the ROI. For Full, metrics \simon{are computed} on the entire image. (LED) models trained on HD pattern outperform (HB) \simon{ones} in all metrics.}
    
    \label{tab:res_globalscene}
    %\vspace{-10pt}
\end{table*}

\begin{figure*}[t]
    \centering
    \includegraphics[width=0.6\linewidth]{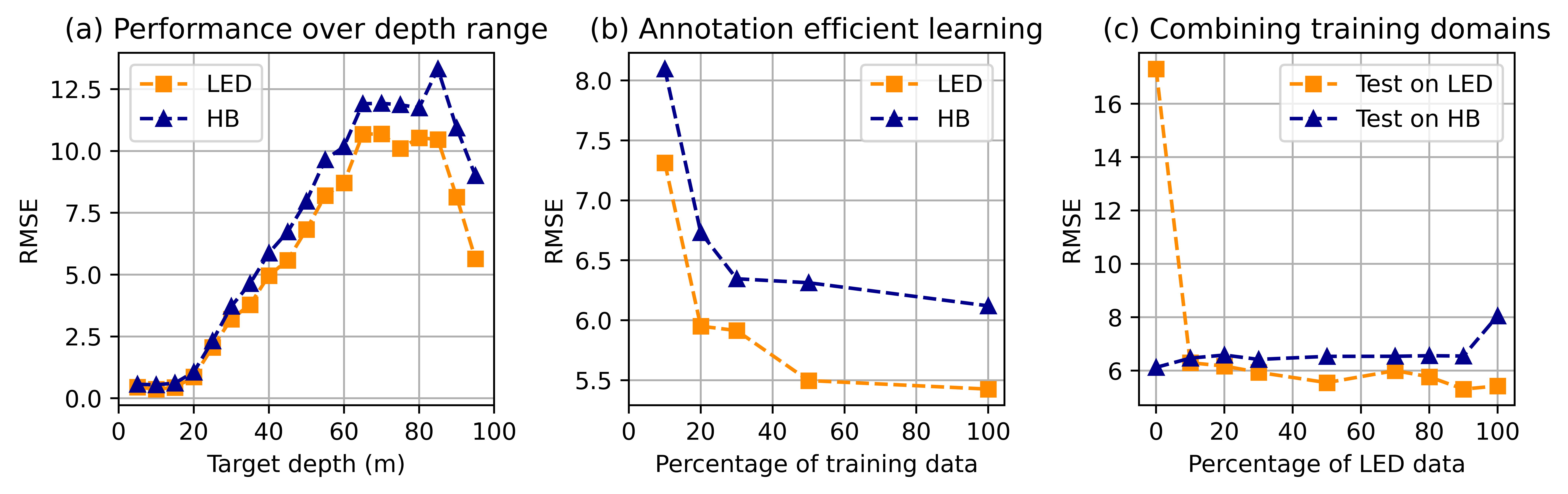}
    \vspace*{-5pt}
    \caption{\textbf{Performance comparison of LED and HB in various settings and metrics on NSDD.} 
    (a) Performance over distance; (b) Performance over training set size; (c) Robustness across domains: 
    The amount of training data is fixed and composed of a given ratio of LED and HB data. Overall, LED achieves better long-range results while being more data-efficient. It also demonstrates improved performance across domains.}
    \label{fig:results_plots}
    \vspace{-15pt}
\end{figure*}

% \vspace{-6pt}
\subsection{HD Pattern Impact}
%\vspace{-3pt}
\label{sec:hd_pattern_impact}

\simon{We assess LED impact on depth estimation by comparing results with and without pattern.}
As reference, we take models trained on high beam (HB), while the variants use our HD pattern (LED). 
Quantitative and qualitative results are %respectively 
presented in \cref{tab:res_HP} and \cref{fig:qualitatif-results}.\\
\textbf{Quantitative Results.}
\simon{We evaluate our method on the encoder-decoder and two SOTA architectures commonly used in daytime scenarios, Adabins \cite{bhat2021adabins} and DepthFormer \cite{li2023depthformer}.}
Their use of transformer architectures offers valuable insights into LED behavior on more recent architectures compared to the convolutional ones that are common in embedded devices.

Comparing the encoder-decoder LED with HB (\cref{tab:res_HP}), we observe a significant improvement in metrics within the region-of-interest (ROI): LED yields -11\% RMSE.
\simon{Using LED, the model provides more accurate distance estimations. 
However, its errors are often related to pattern occlusions, which affect both near and far objects, resulting in a degradation of relative metrics (Abs Rel, Sq Rel).}
All metrics assessing global precision \simon{are improved.}

Compared to the SOTA approaches \cite{bhat2021adabins,li2023depthformer}, LED-trained encoder-decoder either outperforms or matches the performance of their \simon{HB models.}
\simon{Given that the encoder-decoder is less tailored for depth estimation than more intricate SOTA designs, this result underscores the substantial improvements yield by LED, demonstrating that using the HD informative pattern even with a straightforward architecture is promising for challenging nighttime scenes.}

With \simon{LED}, Adabins demonstrates a substantial improvement compared to the HB model\simon{:} -24.06\% RMSE and -6.70\% Abs Rel. Conversely, DepthFormer showcases -8.00\% in RMSE but +5.28\% in Abs Rel. Note that the most significant enhancement is observed in Adabins, \simon{which have the worst results by night on HB.} \\
Improvements seen across diverse architectures confirm LED agnosticity to architectures. It implies potential effectiveness with future methods. \\
\textbf{Performance Over Distance.} 
\simon{Limited nighttime visibility makes depth estimation of distant objects challenging.}
In \cref{fig:results_plots} (a), we \simon{show} the performance of the encoder-decoder \simon{over distances.}
\simon{LED maintains performance for close object and exhibits greater enhancement at longer ranges, thus addressing depth estimation of distant objects at night.} \\
%\vspace{-7pt}
\begin{wraptable}{l}{0.45\textwidth}{
\vspace{-13pt}
\centering
\resizebox{0.30\columnwidth}{!}{
    \begin{tabular}{c | c c c} \toprule  
    Pattern & RMSE$\downarrow$ & Abs Rel$\downarrow$ & SILog$\downarrow$ \\ \midrule
    LED & \textbf{5.497} & 0.209 &  \textbf{12.635}\\ \midrule
    HB & 6.443 & 0.095 & 13.723\\  \midrule
    HL & 6.668 & \textbf{0.068} & 14.308\\  \midrule
    VL & 7.360 & 0.172 & 15.774\\  \bottomrule
    \end{tabular}
}
\caption{\textbf{Performances of the encoder-decoder using various patterns on NSDD.} VL and HL refers to vertical and horizontal lines. Results are obtained from 50\% of NSDD training data.}
\label{tab:pattern-study}
}
\vspace{-10pt}
\end{wraptable} 
\textbf{Pattern Study.} 
\simon{We generated additional data with various illumination patterns to compare the performance of the LED checkerboard against other common structured light patterns (horizontal and vertical lines). \Cref{tab:pattern-study} shows that the checkerboard pattern outperforms the others across most metrics. While horizontal lines show slight improvement in relative metrics, they cause a significant decline in others.}
\begin{table*}[t]
\setlength{\tabcolsep}{8pt} % Default value: 6pt
\renewcommand{\arraystretch}{1.25} % Default value: 1
    \centering
\resizebox{0.65\textwidth}{!}{
    \begin{tabular}{c | c c c c c c c c c c} 
    % \hline 
    \toprule
         Domain &  RMSE $\downarrow$&  Abs Rel $\downarrow$&  $\text{Log}_{10}$ $\downarrow$&  RMSE Log $\downarrow$&  SILog $\downarrow$&  Sq Rel $\downarrow$&  $\delta^1$ $\uparrow$&  $\delta^2$ $\uparrow$& $\delta^3$ $\uparrow$\\ 
         % \hline \hline 
         \midrule
         (HB$\rightarrow$HB)&  \textbf{6.1204}&  \textbf{0.0903}&  \textbf{0.0233}&  \textbf{0.1353}&  \textbf{13.4847}&  \textbf{3.3579}&  \textbf{0.9489}&  \textbf{0.9812}& \textbf{0.9908}\\ 
         (HB$\rightarrow$LED)&  17.3113&  0.4479&  0.1955&  0.6817&  52.5132&  20.0263&  0.4699&  0.6652& 0.7658\\ 
         % \hline
         \midrule
         (LED$\rightarrow$LED)&  \textbf{5.4259}&  0.1996&  \textbf{0.0188}&  \textbf{0.1253}&  \textbf{12.4900}&  16.1224&  \textbf{0.9603}&  \textbf{0.9846}& \textbf{0.9927}\\ 
         (LED$\rightarrow$HB)&  8.0537&  \textbf{0.1511}&  0.0338&  0.2026&  20.1120&  \textbf{8.1999}&  0.9158&  0.9626& 0.9796\\
        % \hline
        \bottomrule
    \end{tabular}
    }
    \caption{\textbf{Encoder-decoder performance across domains of NSDD.} We denote (training domain $\rightarrow$ testing domain). 
    The model trained on (HB) fails entirely when tested on (LED). Yet, (LED)-trained models \simon{are able} to estimate depth when tested on the (HB) domain.} 
    \label{tab:res_across_domain}
    %\vspace{-8pt}
\end{table*}
\begin{table*}[t]
\centering
\resizebox{0.65\linewidth}{!}{
    \begin{tabular}{l | c  c c c c c c c c c} \toprule 
         Pattern & \# Images &  RMSE $\downarrow$&  Abs Rel $\downarrow$&  $\text{Log}_{10}$ $\downarrow$&  RMSE Log $\downarrow$&  SILog $\downarrow$&  Sq Rel $\downarrow$&  $\delta^1$ $\uparrow$&  $\delta^2$ $\uparrow$& $\delta^3$ $\uparrow$\\ 
         % \hline \hline 
         \midrule
         HB & \multirow{2}{*}{Zero-shot} &  19.058 & 0.430 & 0.243 & 0.587 & 27.004 & 8.134 & 0.086 & 0.186 & 0.785 \\ 
         LED &  &  \textbf{18.023} &  \textbf{0.420} & \textbf{0.232} & \textbf{0.563} & \textbf{25.500} & \textbf{7.779} & \textbf{0.096} & \textbf{0.214 }& \textbf{0.867} \\ 
         \midrule
         HB & \multirow{2}{*}{100} &  6.499 & 0.075 & 0.033 & 0.133 & 12.894 & 0.992 &  0.929 & \textbf{0.977} & \textbf{0.991} \\  
         LED &  & \textbf{6.198} & \textbf{0.073} & \textbf{0.032} & \textbf{0.131} & \textbf{12.182} & \textbf{0.898} & \textbf{0.933} & \textbf{0.977} & 0.991 \\ 
        \midrule
         HB & \multirow{2}{*}{500} &  5.447 &  0.058 &  \textbf{0.025} & 0.112 & 10.629 & 0.688 &  0.950 & 0.985 & \textbf{0.995} \\ 
         LED & & \textbf{5.225} & \textbf{0.057} & \textbf{ 0.025} &\textbf{ 0.108} & \textbf{10.038} & \textbf{0.648} &  \textbf{0.953} & \textbf{0.986} & \textbf{0.995} \\ 
        \midrule
         HB & \multirow{2}{*}{1000} &  5.124 &  0.058 &  0.026 & 0.107 & 10.038 & 0.607 &  0.953 & 0.985 & 0.995 \\ 
         LED &  &  \textbf{4.328} &  \textbf{0.044} &  \textbf{0.020} & \textbf{0.089} & \textbf{8.374 }& \textbf{0.430} &  \textbf{0.962} &\textbf{ 0.991} & \textbf{0.997} \\ 
         \bottomrule
    \end{tabular}
    }
    \caption{\textbf{Depth Anything V2 performances on NSDD.} The model fails when used as zero-shot, but \simon{it} correctly estimates depth with few-shot learning. \simon{LED always outperforms HB.}
    }
    \label{tab:depth_any_synth}
    \vspace{-15pt}
\end{table*}

%\vspace{-18pt}
\subsection{Improving Global Scene Understanding}
%\vspace{-3pt}

\simon{We investigate the impact of LED on depth estimation beyond the ROI, highlighting its contribution to a more holistic scene understanding. Results are presented in \cref{tab:res_globalscene}. \\}
\simon{Examining depth estimation both outside the ROI and across the entire image, we observe a consistent performance improvement: -2.66\% RMSE outside the ROI and -3.05\% RMSE on the entire image. We also report enhancement in relative metrics: Abs Rel (resp. -3.03\% and -1.50\%) and Sq Rel (resp. -23.27\% and -9.23\%). 
These results suggest that the HD headlight pattern provides valuable information, such as object size and scale, leading to enhanced overall scene understanding and more accurate depth estimation.
}

%\vspace{-12pt}
\subsection{Annotation-efficient Learning}
%\vspace{-3pt}
Collecting annotated nighttime data is challenging and costly. 
\simon{We explore} whether pattern extra guidance \simon{ can reduce the need} for training data (refer to \cref{fig:results_plots} (b)). 
\simon{Our results show that encoder-decoder model trained with less than 20\% of the LED data outperforms one trained on the full HB dataset, reaching near-peak performance with just 50\%.}
In addition, we observe that incorporating just 10\% of LED data into the HB training set enable the network to learn relevant features and enhance its performances when used with pattern (see \cref{fig:results_plots} (c)). Therefore, any vehicle equipped with HD headlights can apply LED by adding only a few pattern images in their training set, thereby reducing the cost of specialized data.

%\vspace{-12pt}
\subsection{Few-shot Learning}
%\vspace{-3pt}
\label{sec:depth-any}
\simon{Foundation models show strong zero-shot performance across various cameras and content types. However, when tested on NSDD, Depth Anything V2 \cite{depth_anything_v2} performs poorly, with an RMSE of $\sim$19\,m (see \cref{tab:depth_any_synth}), highlighting the complexity of nighttime data and the need for specialized models. It performs slightly better on LED images, possibly due to its ability to leverage the pattern's geometric cues. Fine-tuning with as little as 100 images is sufficient for adapting the network to our distribution and produce consistent depth predictions on both LED and HB.}
LED-models demonstrate significant improvements across all metrics (-15.5\% RMSE with 1,000 images). This underlines the effectiveness of our method in enhancing nighttime depth estimation even on foundation models. \simon{Results on our real-world dataset are provided in the supplementary material.}

\begin{table*}[t]
\centering
\resizebox{0.65\linewidth}{!}{
    
    \begin{tabular}{l | c c c c c c c c c} \toprule 

         Pattern &  RMSE $\downarrow$&  Abs Rel $\downarrow$&  $\text{Log}_{10}$ $\downarrow$&  RMSE Log $\downarrow$&  SILog $\downarrow$&  Sq Rel $\downarrow$&  $\delta^1$ $\uparrow$&  $\delta^2$ $\uparrow$& $\delta^3$ $\uparrow$\\ 
         % \hline \hline 
         \midrule
         \multicolumn{10}{l}{\cellcolor{gray!20}\emph{Encoder-decoder}} \\
         LB&  14.199 &  0.187 &  0.072 &  0.292 &  28.938 &  5.620 &  0.772 &  0.909 & 0.956 \\ 
         LED 0.5°&  11.089 &  0.119 &  0.050 &  0.213 &  21.251 &  2.304 &  0.864 &  0.950 & 0.977 \\ 
         LED 0.25°&  \textbf{8.695} &  0.109 &  \textbf{0.040} &  \textbf{0.190} &  \textbf{18.635} &  \textbf{2.128} &  0.899 &  0.963 & \textbf{0.984} \\ 
         LED 0.125°&  10.154 &  \textbf{0.096} &  \textbf{0.040} &  \textbf{0.190} &  18.862 &  2.132 &  \textbf{0.907} &  \textbf{0.969} & \textbf{0.984} \\ 
         \midrule
         \multicolumn{10}{l}{\cellcolor{gray!20}\emph{DepthFormer}} \\
         LB&  8.777 &  0.138 &  0.050&  0.186 &  17.170 &  3.365 &  0.864 &  0.957 & 0.983 \\ 
         LED 0.5°&  6.810 &  0.101 &  0.041&  0.152 &  14.380 &  1.376 &  0.890 &  0.969 & 0.989 \\ 
         LED 0.25°&  \textbf{5.621} &  0.082 &  \textbf{0.030}&  0.126 &  11.735 &  \textbf{1.163} &  0.923 &  0.978 & 0.992 \\ 
         LED 0.125°&  5.727 &  \textbf{0.076} &  \textbf{0.030}&  \textbf{0.116} &  \textbf{10.754} &  1.261 &  \textbf{0.940} &  \textbf{0.987} & \textbf{0.995} \\ 
         \midrule
         \multicolumn{10}{l}{\cellcolor{gray!20}\emph{DepthAnythingV2 - 1000 training images}} \\
         LB&   7.637 &  0.104 &  0.044&  0.160 &  14.617 &  1.401 &  0.891 &  0.972 & 0.990 \\ 
         LED 0.5°  &  7.017 &  0.109 &  0.047&  0.161 &  14.044 &  1.260 &  0.877 &  0.9697 & 0.989 \\ 
         LED 0.25°  &  6.304 &  0.102 &  0.044 & 0.146 & 12.300 & 1.079 &  0.885 & 0.973 & \textbf{0.993} \\ 
         LED 0.125° &  \textbf{5.785} &  \textbf{0.082} &  \textbf{0.037} & \textbf{0.126} & \textbf{10.673} & \textbf{0.840} &  \textbf{0.921} & \textbf{0.981} & 0.992 \\ 
         % \hline
         \bottomrule
    \end{tabular}
    }
    
    \caption{\textbf{Performance comparison on real-world data using various pattern resolutions.} LED models outperform LB \simon{on} all metrics, \simon{showing} LED \simon{suitability} on real-world scenarios. }
    \label{tab:real_world}
    % \vspace{-15pt}
\end{table*}

\vspace{-6pt}
\subsection{Robustness Across Domains}
%\vspace{-3pt}
\simon{We assess model robustness when operating beyond the original training domain by testing a HB-trained model on pattern data and \textit{vice versa}. We report results in \cref{tab:res_across_domain}. The HB-trained model fails significantly on pattern data (+182.85\% RMSE), while the LED-trained model performs reasonably on HB, with a +32.63\% RMSE. The LED$\rightarrow$HB performance is $\sim$8\,m RMSE, compared to $\sim$17\,m RMSE for HB$\rightarrow$LED.}
\simon{Furthermore, we show that incorporating only 10\% of HB data reduces performance drop (see \cref{fig:results_plots} (c)), limiting RMSE increase to just +17.11\%. It enables a single network to perform well across both domains. This adaptability broadens the method’s applications, allowing selective use of the pattern for higher precision, its absence to avoid glare, or focused pattern projection on specific regions, such as objects of interest}

\vspace{-6pt}
\subsection{Real-world Scenarios}
\begin{wrapfigure}{l}{0.5\columnwidth}
    \vspace{-0pt}
    \centering
     \includegraphics[width=0.5\columnwidth]{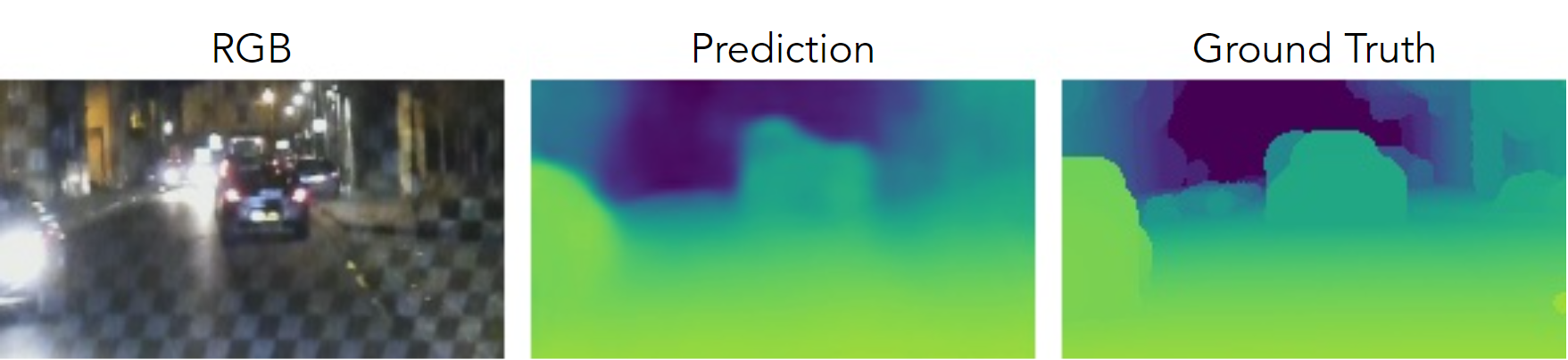}
     %\vspace*{-20pt}
     \caption{\textbf{Real-world qualitative results.} DepthFormer LED produces accurate depth %estimation 
     on complex scenes, even in presence of interfering lights, \eg, car headlights, streetlights.}
     \label{fig:quali_real}
     \vspace{-6pt}
\end{wrapfigure}

%\vspace{-5pt}
We show the applicability of our method in real world scenarios using our in-house dataset. We report results in \cref{tab:real_world}. 
LED significantly boosts performance over LB across all metrics. DepthFormer (0.5°) shows -22.4\% RMSE improvement. The qualitative results in \cref{fig:quali_real} show LED's robustness under interfering light sources, \eg, car headlights, street\,lights. 
\
%\vspace{-0pt}

To account for objects passing through the pattern illumination, we investigate the impact of smaller checkerboard cells. 
Using a size of 0.25°, we observe great enhancement across all metrics (-17.4\% RMSE, -18.8\% Abs Rel against 0.5°). Further increasing the resolution to 0.125° does not lead to significant improvements. \\
In contrast, Depth Anything V2 achieves its best performance with a 0.125° resolution, as it is better suited to capturing fine details. 
When fine-tuned with only 1,000 images, we observe -24.3\% RMSE compared to the LB baseline.
Performances shown by DepthFormer and Depth Anything V2 % (LED)
underpin the benefits of LED in complex real-world scenarios. 

%% file: sections/limitations_conclusion.tex
%\vspace{-15pt}
\section{Limitations and Future Works}
%\vspace{-6pt}
\simon{
The method's reliance on a single reference pattern ensures agnosticism but limits its flexibility. Designing specialized architectures that treat the pattern as an input could enable dynamic pattern optimization and improve generalization.
}

\simon{We tested varying projector-camera distances (real:\,70\,cm,\,NSDD:\,150\,cm), showing LED robustness on similar car models. Future work will assess its usage with larger variations.}

\simon{LED patterns may cause glare for other road users. Leveraging \cite{fleet_programmable_2014} to mask them could mitigate this. Future work should assess both safety concerns and the impact on performance.}

%\vspace{-15pt}
\section{Conclusion}
%\vspace{-6pt}
\simon{We introduce LED, a method for enhancing nighttime depth estimation by leveraging high-definition light patterns projected by modern vehicle headlights. Through extensive experiments, we demonstrate significant improvements in depth perception, both within and beyond illuminated areas. Our method's versatility is highlighted by its successful integration with two state-of-the-art architectures, Adabins and DepthFormer, as well as the foundation model Depth Anything V2. Moreover, LED shows promising real-world performance. We also release the Nighttime Synthetic Drive Dataset, comprising 49,995 fully annotated images. We hope it will serve as a valuable resource for the research community, supporting exploration of various nighttime perception tasks.}

\section*{Acknowledgment}
We would like to thank Amandine Brunetto for her help and insights. We also thank Naimeric Villafruela from the Fablab of Mines Paris for his assistance in creating the real-world prototype. In addition, we extend our warm thanks to the Exwayz team for the excellent collaboration on the real-world ground truth processing. This work was granted access to the HPC resources of IDRIS under the allocation 2025-AD011015334 made by GENCI.

%% file: supplementary.tex
\begin{comment}

\documentclass{bmvc2k}

\usepackage[dvipsnames]{xcolor}
\definecolor{cvprblue}{rgb}{0.21,0.49,0.74}
\hypersetup{allcolors=cvprblue}

\newcommand{\blue}[1]{{\color{cvprblue}#1}}
\newcommand{\red}[1]{{\color{red}#1}}
\newcommand{\todo}[1]{{\color{red}#1}}
\newcommand{\TODO}[1]{\textbf{\color{red}[TODO: #1]}}

\newcommand{\andreic}[1]{\andrei{[\textbf{Andrei}: \textit{#1}]}}
\newcommand{\andrei}[1]{\textcolor{teal}{#1}}
\newcommand{\simon}[1]{\textcolor{olive}{#1}}
\newcommand{\simonc}[1]{\textcolor{olive}{[\textbf{Simon}: \em #1]}}

%for submission
% \newcommand{\andreic}[1]{}
% \newcommand{\andrei}[1]{#1}
% \newcommand{\simonc}[1]{}
% \newcommand{\simon}[1]{#1}

\usepackage{cleveref}
\usepackage{comment}
\usepackage{booktabs}
\usepackage[table]{xcolor}
\usepackage{multirow} 
\usepackage{wrapfig}
%\usepackage{wraptab}

\usepackage{pifont}% http://ctan.org/pkg/pifont
\newcommand{\checkmark}{\ding{51}}%
\newcommand{\xmark}{\ding{55}}%
\usepackage{amssymb}

%\usepackage{dblfloatfix}
\end{comment}

%% Enter your paper number here for the review copy
%\bmvcreviewcopy{1096}

\title{Supplementary Material \\ LED: Light Enhanced Depth Estimation at Night}
\addauthor{Simon de Moreau}{https://simon.demoreau.fr/}{1,2}
\addauthor{Yasser Almehio}{yasser.almehio@valeo.com}{2}
\addauthor{Andrei Bursuc}{https://abursuc.github.io/}{3}
\addauthor{Hafid El-Idrissi}{hafid.el-idrissi@valeo.com}{2}
\addauthor{Bogdan Stanciulescu}{bogdan.stanciulescu@minesparis.psl.eu}{1}
\addauthor{Fabien Moutarde}{https://people.minesparis.psl.eu/fabien.moutarde/}{1}

\addinstitution{
 Mines Paris - PSL University\\
 Paris, France
}
\addinstitution{
 Valeo\\
 Paris, France
}
\addinstitution{
 Valeo AI\\
 Paris, France
}

\runninghead{de Moreau \etal}{LED: Light Enhanced Depth Estimation at Night}

% Any macro definitions you would like to include
% These are not defined in the style file, because they don't begin
% with \bmva, so they might conflict with the user's own macros.
% The \bmvaOneDot macro adds a full stop unless there is one in the
% text already.
\def\eg{\emph{e.g}\bmvaOneDot}
\def\Eg{\emph{E.g}\bmvaOneDot}
\def\etal{\emph{et al}\bmvaOneDot}

%-------------------------------------------------------------------------
% Document starts here
%\begin{document}

\maketitle

\setcounter{section}{0}
\setcounter{figure}{0}
\setcounter{table}{0}
\renewcommand\thesection{\Alph{section}}
\renewcommand\thefigure{S\arabic{figure}}    
\renewcommand\thetable{S\arabic{table}}    

\section{Safety and Regulation}
While using readily available hardware, LED is a research project not meant to be deployed on cars right away. Future works should assess potential safety issues. To maximize their safety, autonomous vehicles needs multiple perception mechanisms that ensure redundancy, compensating for blind spots of other sensors on the car, e.g., LiDAR in rain conditions. One can imagine that this light pattern could be turned off in crowded areas, or when detecting an incoming car after it was initially detected at a longer distance, ensuring safety. 
Regarding regulation, HD headlight technology is novel and the European regulation has authorized just recently the projection of specific HD pattern onto the road. Thus, laws are moving in this direction and evolving with the technology.

\section{Encoder-Decoder Details}

% \section{Encoder-Decoder Details}
\subsection{Encoder-Decoder Architecture}
\begin{figure*}[!t]
    \centering
    \includegraphics[width=0.8\textwidth]{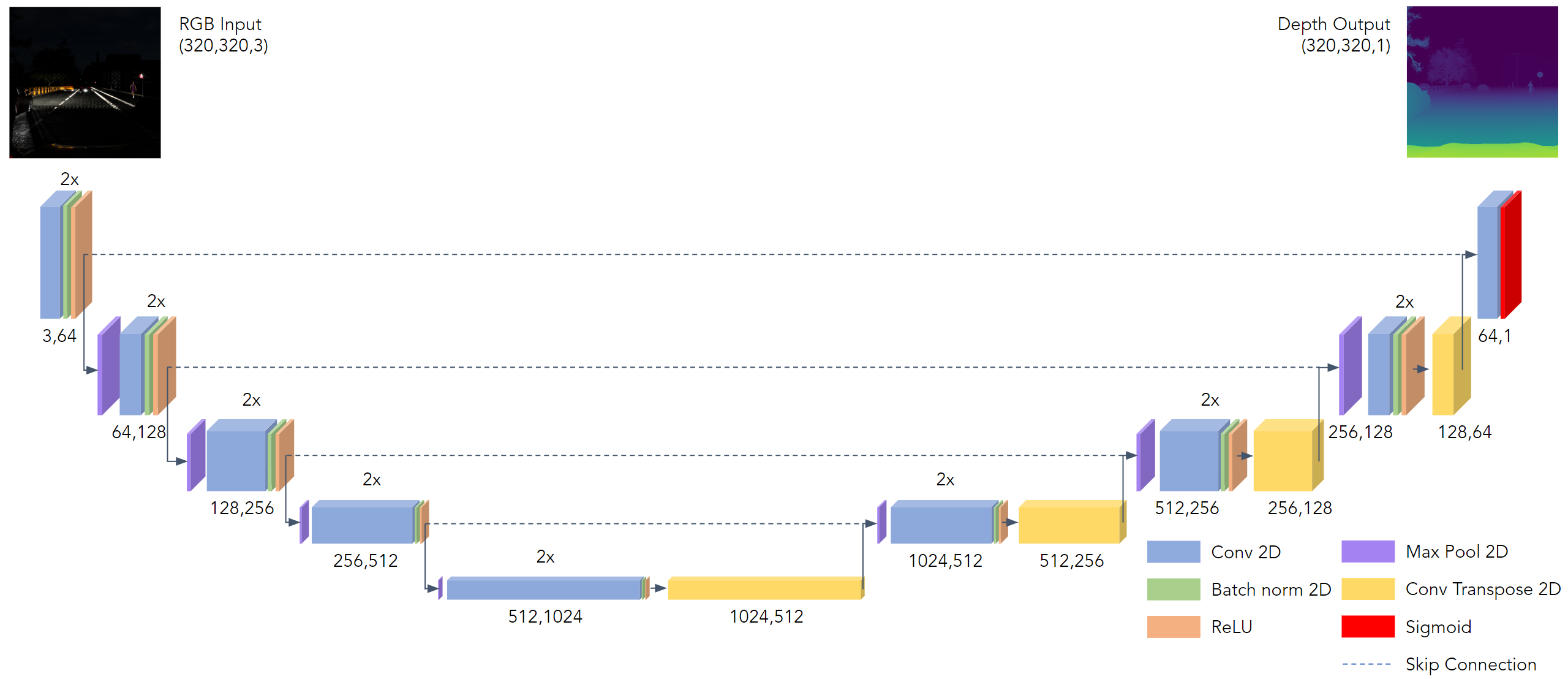}
    \caption{Detailed encoder-decoder architecture.}
    \label{fig:full-archi}
\end{figure*}

LED uses a single pattern implicitly learned by the model, making it architecture-agnostic. This characteristic is demonstrated in the main paper (see section \cref{sec:hd_pattern_impact}) by applying LED to multiple state-of-the-art architectures \cite{bhat2021adabins,li2023depthformer,depth_anything_v2}.
To prove our concept and conduct experiments, we opt for an encoder-decoder architecture with skip connections \cite{U-Net}. The detailed architecture is shown in \cref{fig:full-archi}. Despite the simplicity of this architecture, we show that LED enables the network to achieve performances comparable to or better than other tested SOTA architectures trained without our method.

\subsection{Learning Objectives}
\label{sec:learning-objective}
To train the encoder-decoder, we adopt a combination of loss functions inspired by \cite{HigherResMap}. The primary loss, $\mathcal{L}_{depth}$, is the Log L1 loss presented in \cref{eq:logl1}. It measures the error between the estimated depth, $d_i$, and the corresponding ground truth, $g_i$. This variant of L1 loss attributes less significance to errors occurring at greater distances. This adjustment aligns with the expectation that given errors, e.g., 1\,m, should have greater impact when within a few meters of the camera but are more tolerable at extended distances:

\begin{equation}
\label{eq:logl1}
    \mathcal{L}_{depth} = \frac{1}{N} \sum_{i=1}^{N} |log(d_i) - log(g_i)|.
\end{equation}

We address edges fidelity, particularly challenging in low-light conditions. To this end, we incorporate a loss, $\mathcal{L}_{grad}$, that specifically emphasizes on gradients. As we believe edges sharpness is important regardless of the distance, we employ a standard L1 Loss expressed in \cref{eq:l1grad} instead of a logarithmic one used in \cite{HigherResMap}. Moreover, our experiments revealed no performance improvements with the Log version. $\nabla_x(d_i)$ and $\nabla_y(d_i)$ respectively represent the spatial derivative of $d_i$ along the x and y-axis,

\begin{equation}
\label{eq:l1grad}
\mathcal{L}_{grad} =  \frac{1}{N} \sum_{i=1}^{N} | \nabla_x(d_i) - \nabla_x(g_i) |   
+ | \nabla_y(d_i) - \nabla_y(g_i) |.
\end{equation}

Similar to \cite{HigherResMap}, we want to ensure accurate surfaces representation in depth maps. 

The depth normals are estimated at each pixel using 
$n_i^a \equiv [ -\nabla_x(a_i),- \nabla_y(a_i), 1]^\intercal$. 
The cosine similarity loss, expressed in \cref{eq:lossnormal}, 
is then employed to compare estimated and ground truth normals. 
$\langle .,.\rangle$ denotes vector inner product operation:

\begin{equation}
\label{eq:lossnormal}
    \mathcal{L}_{normal} = \frac{1}{N} \sum_{i=1}^{N} | 1 - \frac{\langle n_i^d,n_i^g\rangle}{\sqrt{n_i^d,n_i^d}\sqrt{n_i^g,n_i^g}} |.
\end{equation}

Finally, our learning objective can be expressed as \cref{eq:totalloss}, we set $\lambda_1=1$ and $\lambda_2=1$:

\begin{equation}
\label{eq:totalloss}
    \mathcal{L} = \mathcal{L}_{depth} + \lambda_1\mathcal{L}_{grad} + \lambda_2\mathcal{L}_{normal}.
\end{equation}

\subsection{Loss Study}
To determine the impact of each individual loss, we conduct an ablation study (see \cref{tab:loss-study}). Both L1 and Log L1 losses are tested for $\mathcal{L}_{depth}$ and $\mathcal{L}_{grad}$. Our findings indicate that each selected loss positively influences performance. The combination outlined in \cref{sec:learning-objective} demonstrates the most favorable trade-off between metrics.

\begin{table*}[t]
\setlength{\tabcolsep}{8pt} % Default value: 6pt
\renewcommand{\arraystretch}{1.25} % Default value: 1
    \centering
\resizebox{0.8\textwidth}{!}{
    
    \begin{tabular}{c  c | c  c | c | c c c c c c c c c c} \toprule 
          \multicolumn{2}{c|}{$\mathcal{L}_{depth}$} & \multicolumn{2}{c|}{$\mathcal{L}_{grad}$} & $\mathcal{L}_{normal}$ & \multicolumn{9}{c}{Metrics} \\ \midrule
          L1 & Log L1 & L1 & Log L1 & &  RMSE $\downarrow$&  Abs Rel $\downarrow$&  $\text{Log}_{10}$ $\downarrow$&  RMSE Log $\downarrow$&  SILog $\downarrow$&  Sq Rel $\downarrow$&  $\delta^1$ $\uparrow$&  $\delta^2$ $\uparrow$& $\delta^3$ $\uparrow$\\ 
         % \hline \hline 
         \midrule
         \checkmark & \xmark & \xmark & \xmark & \xmark & 5.979 & \textit{0.191} & \underline{0.020} & \textit{0.132} & 13.032 & \underline{14.352} & \textit{0.957} & \textit{0.982} & \textit{0.991}\\ \midrule
         \xmark & \checkmark & \xmark & \xmark & \xmark & 5.852 & \underline{0.187} & \textit{0.021} & \textit{0.132} & 13.122 & \textbf{12.984} & \textit{0.957} & \underline{0.983} & \underline{0.992}\\ \midrule
         \xmark & \checkmark & \checkmark & \xmark & \xmark & 5.760 & 0.201 & \underline{0.020} & 0.137 & 13.616 & 16.174 & \underline{0.958} & \underline{0.983} & \underline{0.992}\\ \midrule
         \xmark & \checkmark & \xmark & \checkmark  & \xmark & \underline{5.386} & 0.202 & \textit{0.021} & \textbf{0.124} & \underline{12.373} & 15.901 & \underline{0.958} & \textbf{0.985} & \textbf{0.993} \\ \midrule
         \xmark & \checkmark & \xmark & \xmark &\checkmark & \textbf{5.309} & 0.206 & 0.022 & 0.149 & 14.862 & 16.048 & 0.954 & 0.981 & 0.990 \\ \midrule
         \xmark & \checkmark & \xmark & \checkmark  &\checkmark & 5.809 & \textbf{0.186} & \textbf{0.018} & \underline{0.125} & \textbf{12.210} & \textit{14.683} & \textbf{0.961} & \textbf{0.985} & \textbf{0.993} \\ \midrule
         \xmark & \checkmark & \checkmark & \xmark  &\checkmark & \textit{5.422} & 0.200 & \textbf{0.018} & \underline{0.125} & \textit{12.503} & 16.120 & \textbf{0.961} & \textbf{0.985} & \textbf{0.993} \\
         
         % \hline
         % \hline
         \bottomrule

    \end{tabular}
    }
    
\caption{\textbf{Loss study. }Performance comparison using several combination of losses: \textbf{1st best}, \underline{2nd best}, \textit{3rd best}.}
\label{tab:loss-study}
    %\vspace{2cm}
    
\end{table*}

\section{Depth Anything V2 Details}
\begin{figure}[t!]
    \centering
    \includegraphics[width=0.5\linewidth]{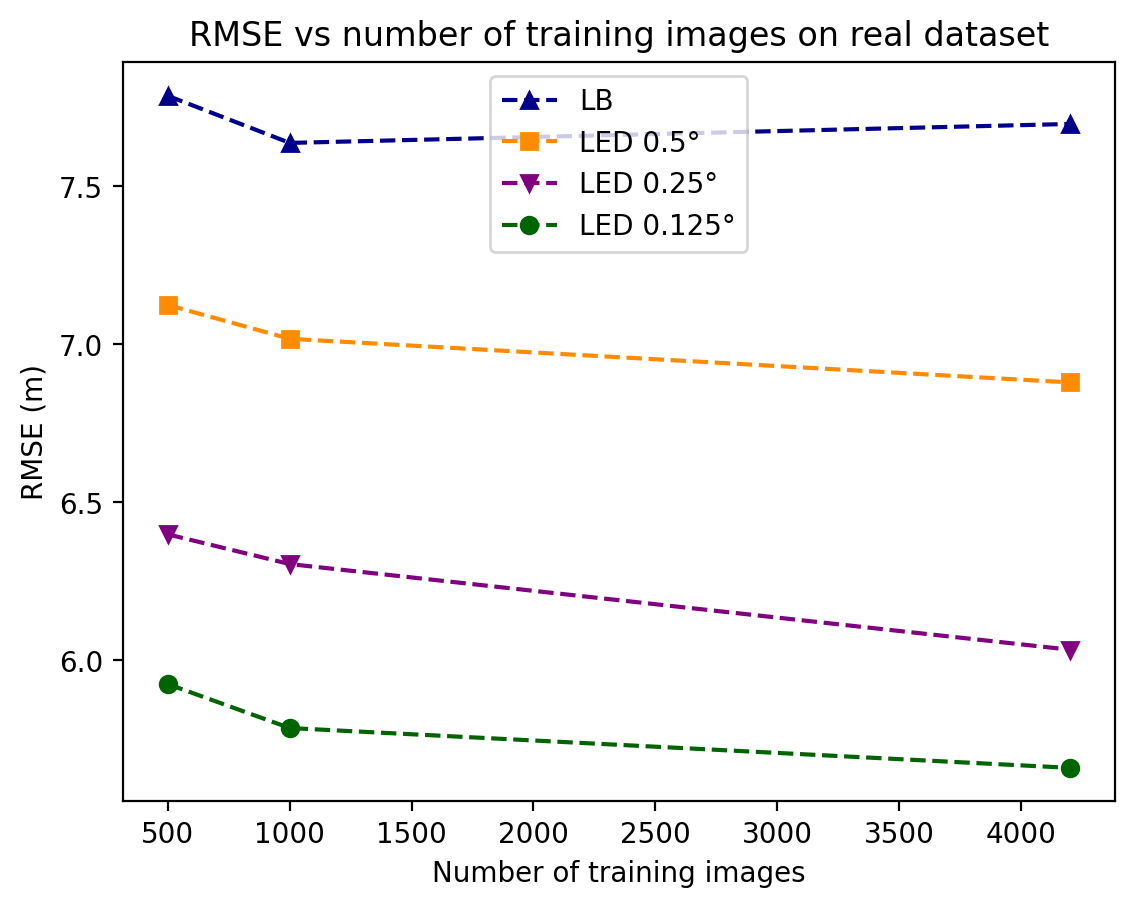}
    \caption{\textbf{Few-shot learning performances of Depth Anything V2 on our real dataset.} All LED patterns demonstrate superior performances over LB. The model is near its maximal performances with only 1,000 training images.}
    \label{fig:ADv2_real_fewshot}
\end{figure}

To fine-tune Depth Anything V2 \cite{depth_anything_v2} on our datasets, we used the metric model pre-trained on KITTI. We follow their indications for specializing the model to metric depth estimation. For each given number of training images, the split was selected randomly in the whole dataset. 
 \Cref{fig:ADv2_real_fewshot} illustrates the RMSE of Depth Anything V2 across different numbers of fine-tuning images. We observe that the model is near its performance peak with just 1,000 images. All LED-enhanced models outperform the LB, also this architecture benefits from finer resolution in the pattern, demonstrating its best performances with LED 0.125°. 

\section{ROI Details}
\subsection{ROI Definition}
To assess the impact of the LED pattern, we define a Region of Interest (ROI) representing the illuminated area in most images and calculate our metrics within this region. We center-crop the image to $640 \times 640$\,px and then resize it to $320 \times 320$\,px. In the resulting image, the ROI consists of pixels with coordinates satisfying: $20 \leq p_x \leq 270$ and $165 \leq p_y \leq 210$, where $p_x$ and $p_y$ are the pixel coordinates along the $x$ and $y$ axes, respectively.

\subsection{ROI-Only Training}
To assess the impact of our method, we focus on the illuminated area. Thus, it is reasonable to evaluate performance when trained exclusively within the ROI. We report results of this experiment in \cref{tab:res_roi-only}. We observe that training solely within the ROI enhances Abs Rel and Sq Rel metrics, although other metrics show a decline. Since the network's ability to estimate depth beyond the ROI is valuable for many applications, we 
do not pursue this approach of ROI-only training.

\begin{table*}[t]
\setlength{\tabcolsep}{8pt} % Default value: 6pt
\renewcommand{\arraystretch}{1.25} % Default value: 1
    \centering
\resizebox{0.7\textwidth}{!}{
    
    \begin{tabular}{c | c c c c c c c c c} \toprule 

         Image input&  RMSE $\downarrow$&  Abs Rel $\downarrow$&  $\text{Log}_{10}$ $\downarrow$&  RMSE Log $\downarrow$&  SILog $\downarrow$&  Sq Rel $\downarrow$&  $\delta^1$ $\uparrow$&  $\delta^2$ $\uparrow$& $\delta^3$ $\uparrow$\\ 
         % \hline \hline 
         \midrule
            Reference &  \textbf{5.4259}&  0.1996&  \textbf{0.0188}&  \textbf{0.1253}&  \textbf{12.4900}&  16.1224&  \textbf{0.9603}&  \textbf{0.9846}& 0.9927\\          
         \midrule
          ROI-Only&  5.6931&  \textbf{0.1145}&  0.0199&  0.1428&  14.2131&  \textbf{4.4541}&  0.9585&  0.9841& \textbf{0.9932}\\ 
         \bottomrule

    \end{tabular}
    }
    
\caption{\textbf{Encoder-decoder performances on NSDD} between ROI-only training and reference from section \cref{sec:hd_pattern_impact}.}
\label{tab:res_roi-only}
    %\vspace{-0.3cm}
    
\end{table*}

\begin{table*}[t]
\setlength{\tabcolsep}{8pt} % Default value: 6pt
\renewcommand{\arraystretch}{1.25} % Default value: 1
    \centering
\resizebox{0.7\textwidth}{!}{
    
    \begin{tabular}{l | c c c c c c c c c} \toprule 

         Resolution&  RMSE $\downarrow$&  Abs Rel $\downarrow$&  $\text{Log}_{10}$ $\downarrow$&  RMSE Log $\downarrow$&  SILog $\downarrow$&  Sq Rel $\downarrow$&  $\delta^1$ $\uparrow$&  $\delta^2$ $\uparrow$& $\delta^3$ $\uparrow$\\ 
         % \hline \hline 
          \midrule
         200\,px&  5.9413&  \textbf{0.1073}&  0.0206&  0.1270&  12.6227&  \textbf{3.9736}&  0.9545&  0.9228& 0.9921\\ 
         \midrule
         320\,px&  \textbf{5.4259}&  0.1996&  0.0188&  \textbf{0.1253}&  \textbf{12.4900}&  16.1224&  0.9603&  \textbf{0.9846}& \textbf{0.9927}\\          
         \midrule
         640\,px&  5.7050&  0.2109&  \textbf{0.0185}&  0.1420&  14.1092&  17.4085&  \textbf{0.9668}&  0.9837& 0.9893\\ 
         \bottomrule

    \end{tabular}
    }
    
\caption{\textbf{Encoder-decoder performances on NSDD} at various resolutions.}
\label{tab:res_px}
    %\vspace{-0.3cm}
    
\end{table*}

\begin{figure*}[b]
    \centering
    \includegraphics[width=0.7\textwidth]{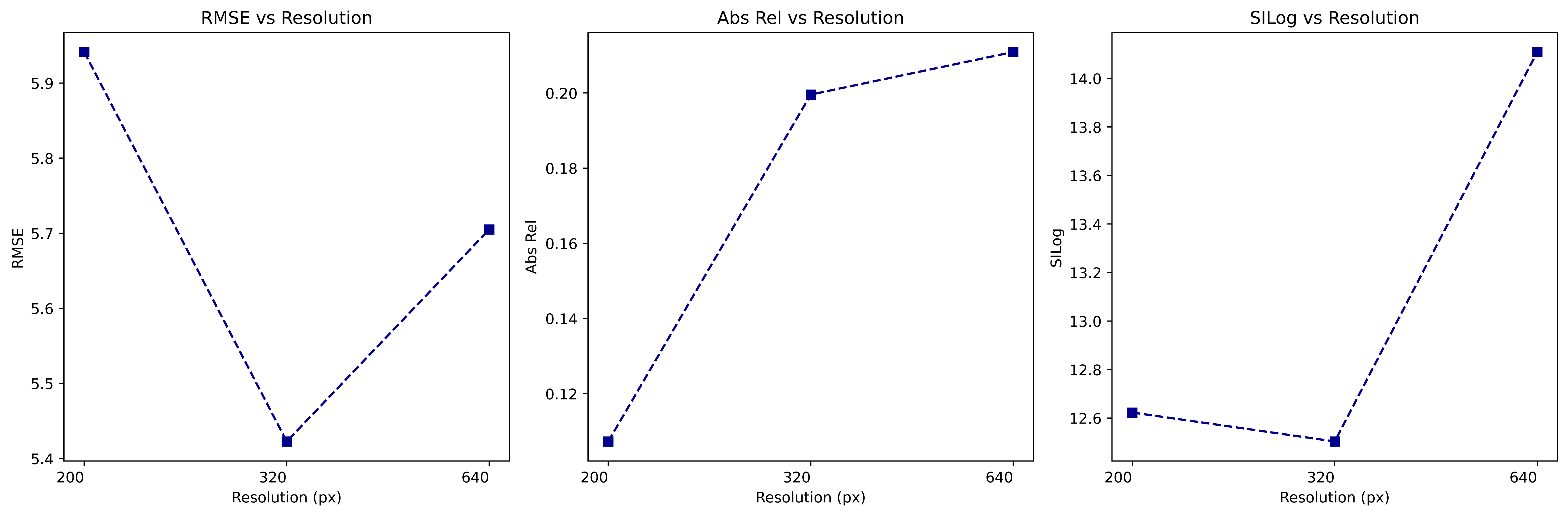}
    \caption{\textbf{Encoder-decoder performances on NSDD} using multiple resolutions}
    \label{fig:perf-reso}
\end{figure*}

\begin{figure}[t]

    \centering
    \includegraphics[width=0.49\textwidth]{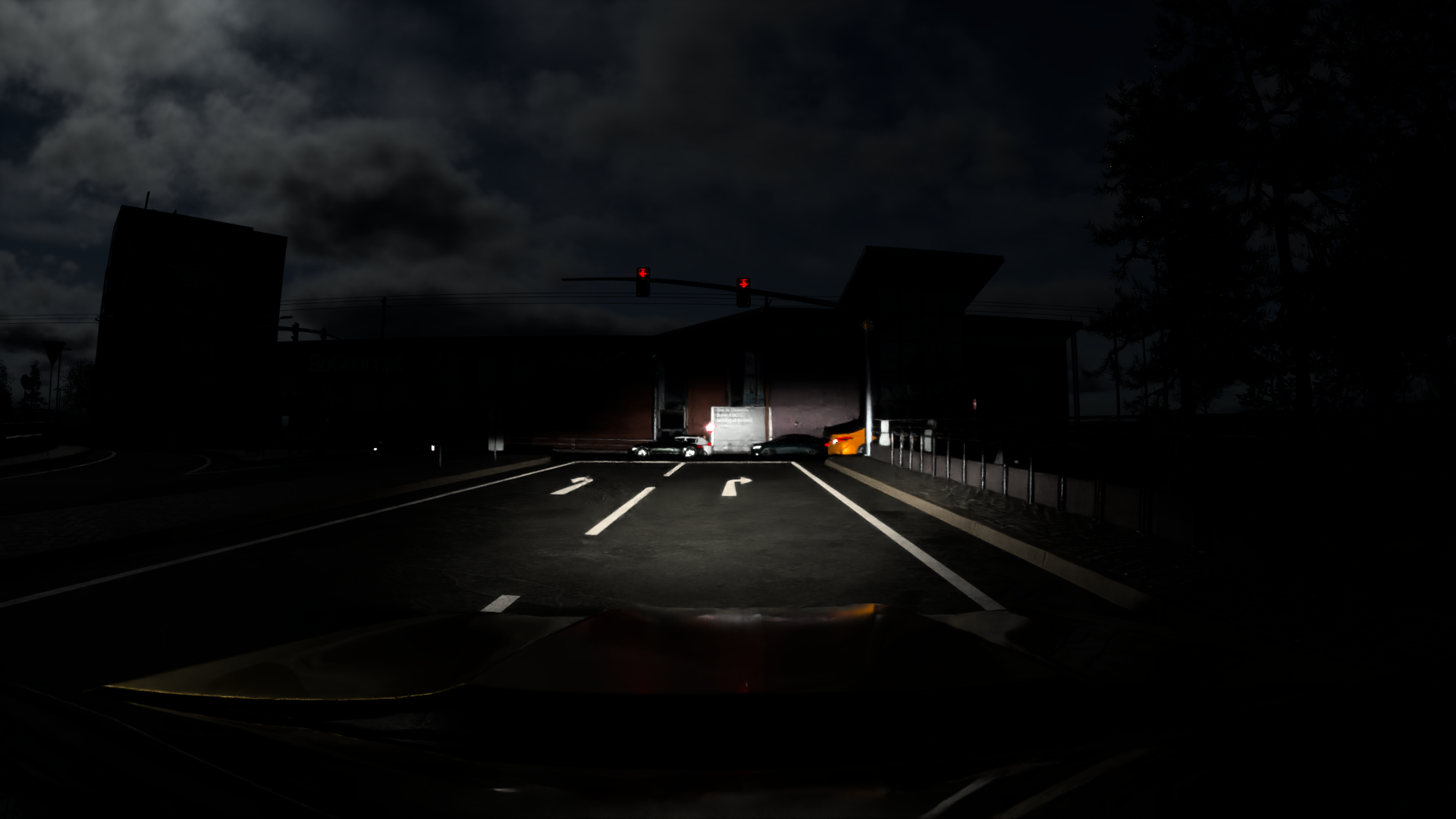}
    \includegraphics[width=0.49\textwidth]{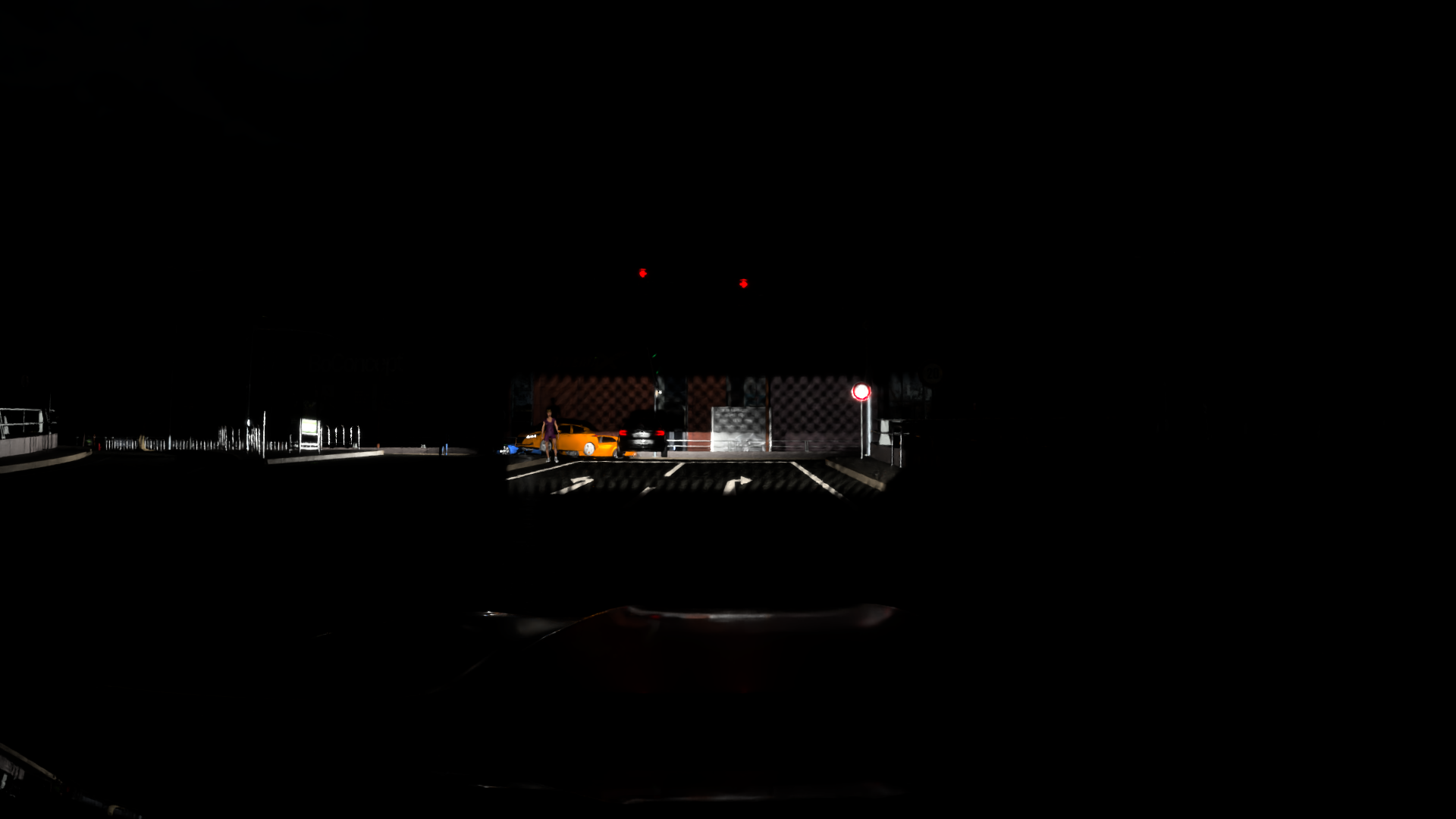}
    \includegraphics[width=0.49\textwidth]{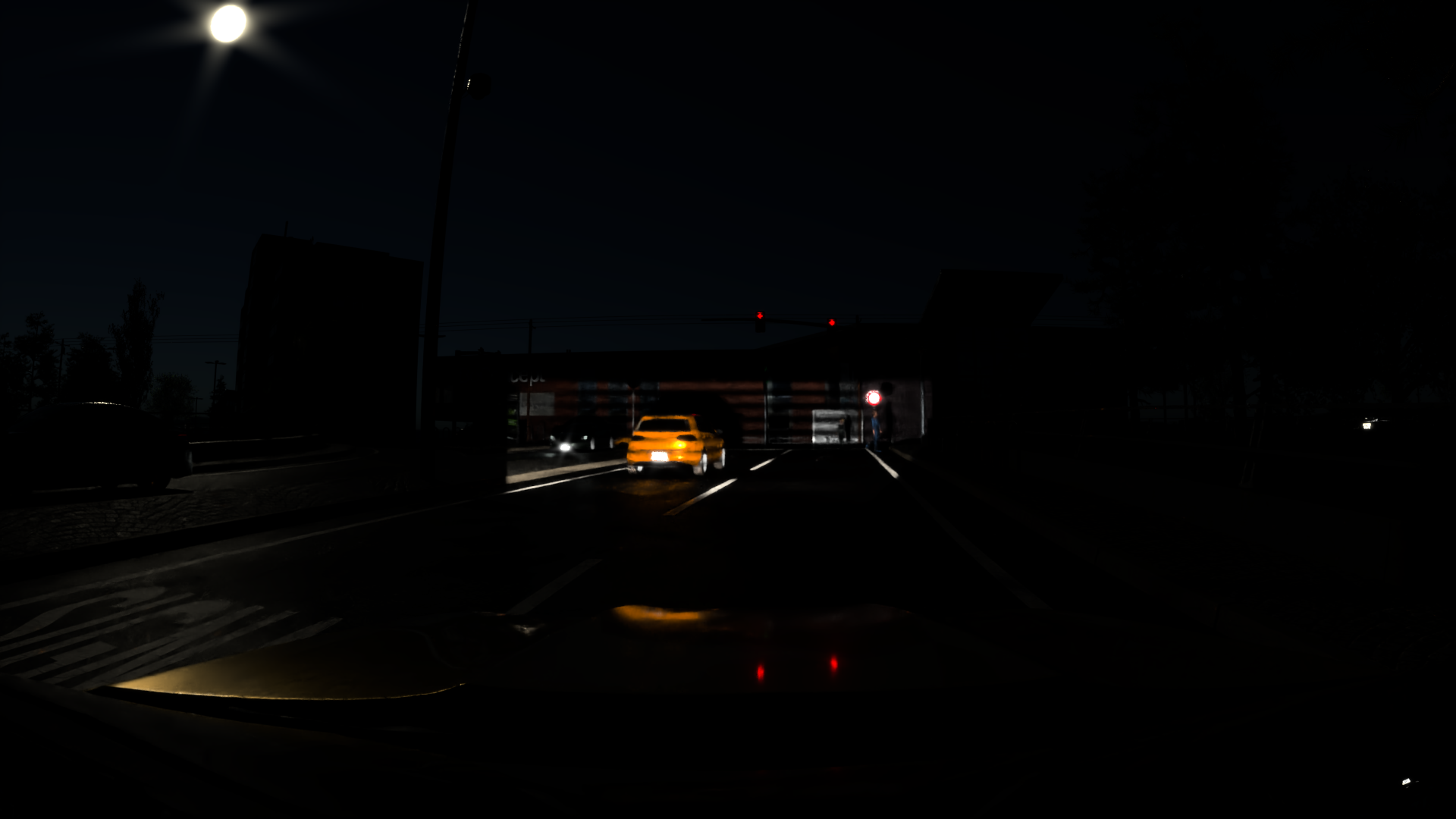}
    \includegraphics[width=0.49\textwidth]{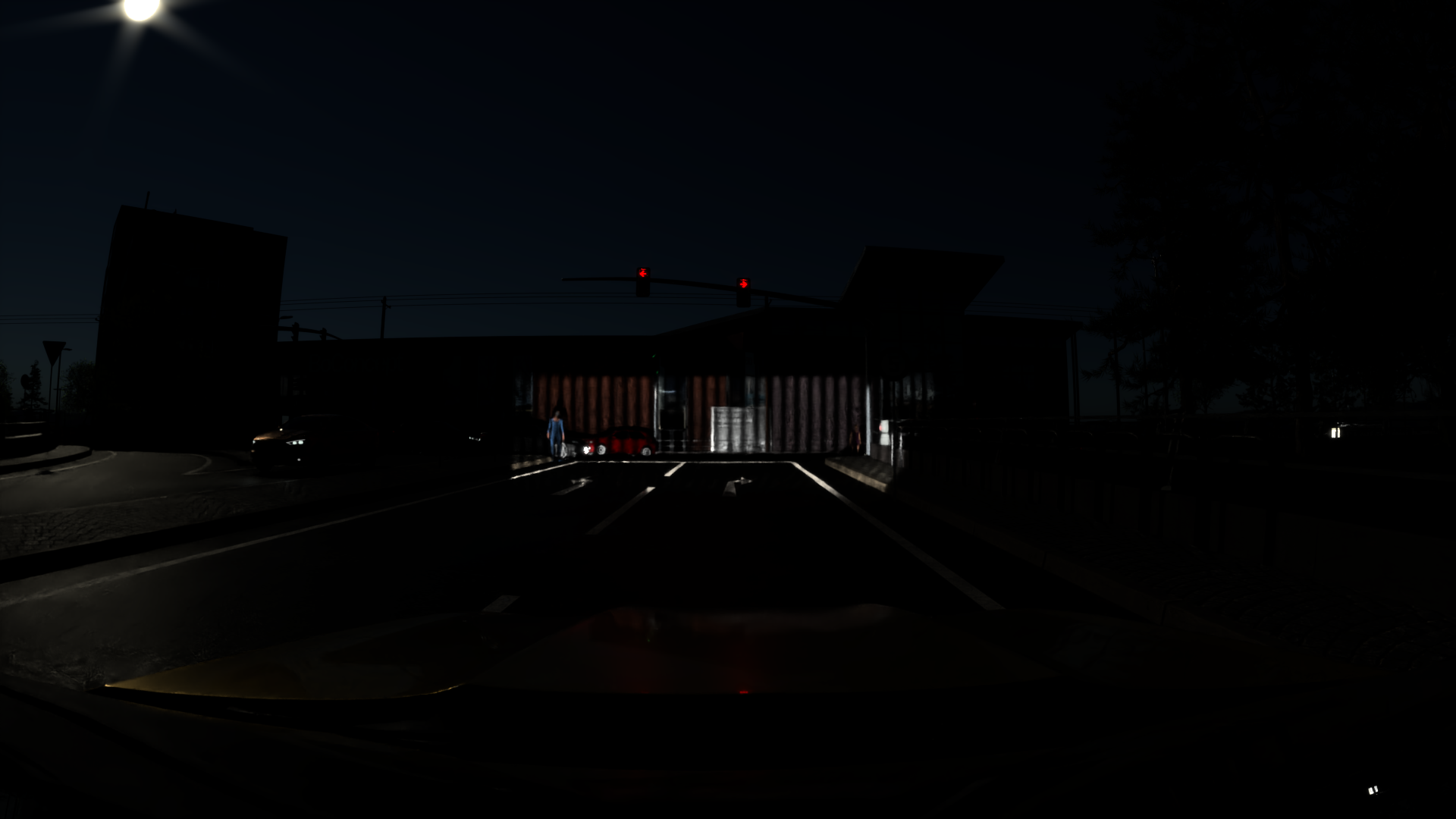}
    \caption{\textbf{Example of illuminations pattern tested}: high-beam (top-left), checkerboard (top-right), horizontal lines (bottom-left) and vertical lines (bottom-right).}
    \label{fig:VL-HL}
\end{figure}

\section{Resolution Impact}
Resolution and performance are usually highly correlated in computer vision, particularly in low-light conditions. To better understand the impact of resolution on our method, we train the encoder-decoder with center-cropped area resized at various resolutions. Results in \cref{tab:res_px} indicate that increasing resolution up to $640 \times 640$\,px does not improve performance. Conversely, decreasing resolution to $200 \times 200$\,px appears to enhance relative metrics while degrading others. Since most valuable cues in our method come from the pattern, these findings suggest that a resolution of $320 \times 320$\,px offers a favorable trade-off for pattern visibility and scene interpretation within our setup (as shown in \cref{fig:perf-reso}).

\begin{table*}[t]
\setlength{\tabcolsep}{8pt} % Default value: 6pt
\renewcommand{\arraystretch}{1.25} % Default value: 1
    \centering
\resizebox{0.7\textwidth}{!}{
    
    \begin{tabular}{c | c c c c c c c c c} \toprule 

         Pattern&  RMSE $\downarrow$&  Abs Rel $\downarrow$&  $\text{Log}_{10}$ $\downarrow$&  RMSE Log $\downarrow$&  SILog $\downarrow$&  Sq Rel $\downarrow$&  $\delta^1$ $\uparrow$&  $\delta^2$ $\uparrow$& $\delta^3$ $\uparrow$\\ 

         \midrule
         LED&  \textbf{5.5179}&  0.1977&  \textbf{0.0196}&  \textbf{0.1233}&  \textbf{12.2798}&  15.6867&  \textbf{0.9593}&  \textbf{0.9849}& \textbf{0.9931}\\ \midrule
         HB&  6.0937&  0.0867&  0.0216&  0.1298&  12.8895&  3.1352&  0.9521&  0.9824& 0.9917\\ \midrule
         VL&  7.3598&  0.1723&  0.0264&  0.1600&  15.7743&  6.5203&  0.9393&  0.9751& 0.9876\\ \midrule
         HL&  6.6679& \textbf{0.0681}&  0.0254&  0.1445&  14.3077&  \textbf{1.2319}&  0.9370&  0.9776& 0.9903\\ 
         
         \bottomrule

    \end{tabular}
    }
    
\caption{\textbf{Comparison of encoder-decoder performances using various patterns}: LED use the checkerboard, HB stands for high-beam, VL and HL are vertical and horizontal lines respectively.}
\label{tab:res_other_patterns}

\end{table*}

\section{Examples of Other Patterns}
To find a suitable pattern for our application, we generate  half-sized datasets featuring commonly used structured light patterns: checkerboard, horizontal, and vertical lines (see \cref{fig:VL-HL}).
Comprehensive metrics from this experiment are available in \cref{tab:res_other_patterns}, the checkerboard pattern demonstrates significantly superior performances.

\section{LED Impact On Other Tasks}

We introduce the LED lighting pattern for improving performance on geometric tasks.
We study its impact on the performance of other tasks that might be running in the same time on the vehicle, taking the example of semantic segmentation. To this end, we train Mask2Former \cite{cheng2021mask2former} using the available annotation in the Nighttime Synthetic Drive Dataset. Results are reported in \cref{tab:semantic}. LED doesn't improve, nor degrade performances. We even note a better stability with less variance over runs. Therefore, the similar performance on both domains suggests that LED enhances geometric tasks with limited impact on semantic ones.

\begin{table}[h!]
    \centering
    \begin{tabular}{l | c} \toprule 

         Pattern&  mIoU $\uparrow$\\%&  mAcc $\uparrow$ \\ 
         % \hline \hline 
         \midrule
                  \multicolumn{2}{l}{\cellcolor{gray!20}\emph{Mask2Former}} \\
         LB&  51.45 $\pm$ 4,00 \\%& 59,83 $\pm$ 4,48\\ 
         LED& 51.49 $\pm$ 0,76 \\%& 57,24 $\pm$ 1,17 \\ 
         \bottomrule

    \end{tabular}
    %}
    \caption{\textbf{Mask2Former \cite{cheng2021mask2former} Semantic segmentation results on NSDD.}}
    \label{tab:semantic}
\end{table}

\begin{table*}[t]
\centering
\resizebox{0.8\textwidth}{!}{
    
    \begin{tabular}{c | c c c c c c c c c} \toprule 

         Test set&  RMSE $\downarrow$&  Abs Rel $\downarrow$&  $\text{Log}_{10}$ $\downarrow$&  RMSE Log $\downarrow$&  SILog $\downarrow$&  Sq Rel $\downarrow$&  $\delta^1$ $\uparrow$&  $\delta^2$ $\uparrow$& $\delta^3$ $\uparrow$\\ 
         % \hline \hline 
         \midrule
         LED (0.5°)&  6.859 &  0.102 &  0.041 &  0.153 &  14.162 &  1.511 &  0.894 &  0.971 & 0.989 \\ 
         LED (0.25°)&  5.787 &  0.082 &  0.033 &  0.126 &  11.875 &  1.0839 &  0.929 &  0.983 & 0.994 \\ 
         LED (0.125°)&  5.730 &  0.077 &  0.030 &  0.116 &  10.913 &  1.149 &  0.958 &  0.989 & 0.995 \\ 
         \bottomrule

    \end{tabular}
    }
     \caption{\textbf{DepthFormer results on real dataset} when trained on 33\% of each checkerboard resolutions (0.5°, 0.25°, 0.125°).}
     
     \label{tab:multiple_res}
\end{table*}

\section{Using Multiple Checkerboard Resolutions}
Any HD headlight can project LED's checkerboard, although with a resolution adjustment. To evaluate its impact on model performances, we propose to train a model with 3 checkerboard resolutions (0.5°, 0.25° and 0.125°) evenly split, and evaluate on each one. 
\Cref{tab:multiple_res} demonstrates that the model is able to perform well on all resolutions although with a minor performance degradation compared to single resolution training. This shows our model's ability to generalize over multiple resolutions, thus, various HD headlights or car models.

\begin{figure*}[h!]

    \centering
     \includegraphics[width=\linewidth]{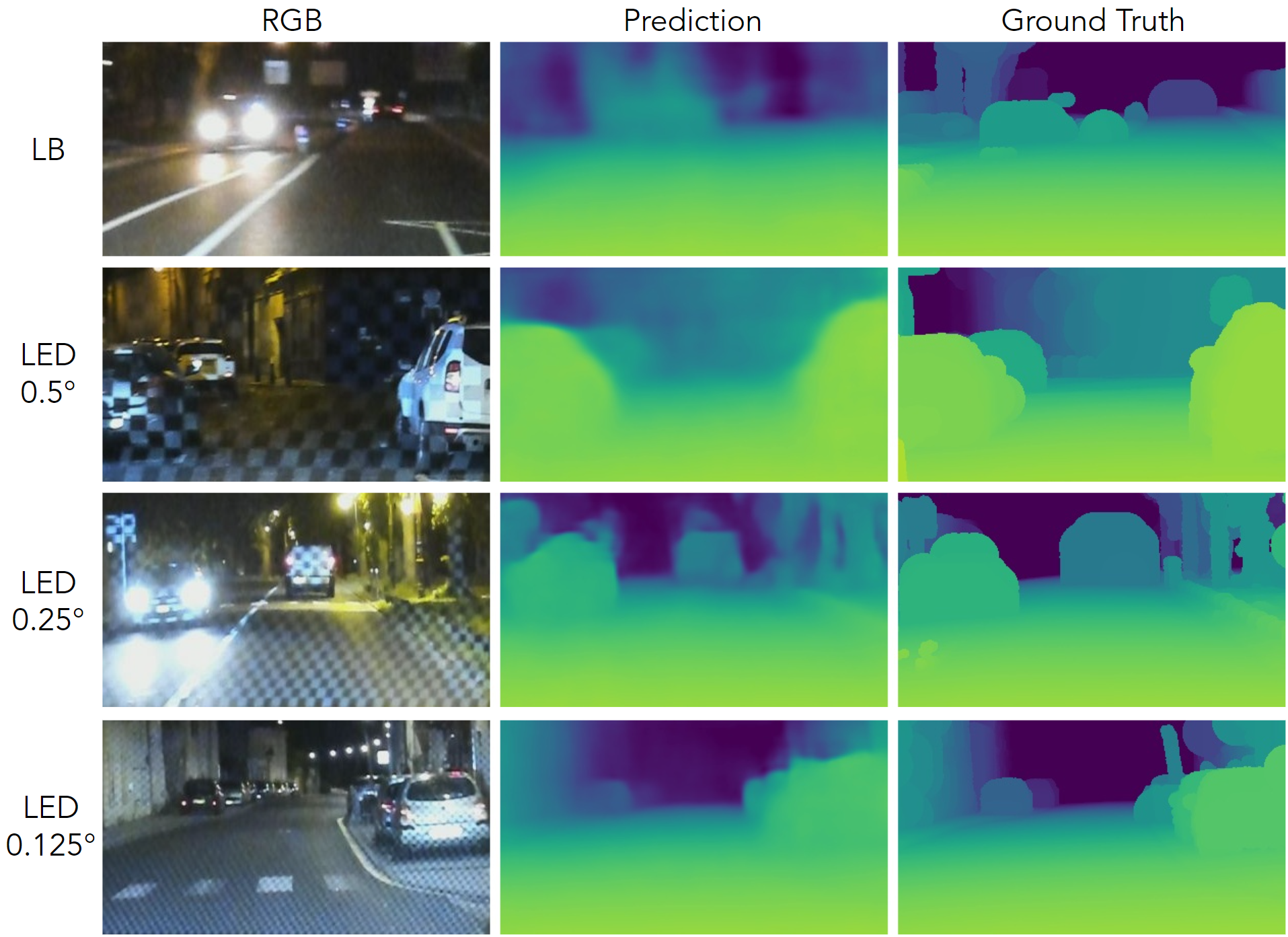}
     \caption{\textbf{Qualitative results on real-world scenarios}. LED-trained models are more accurate and object edges are better defined.}
     \label{fig:sup_quali_real}
     %\vspace{-15pt}
\end{figure*}
\newpage
\section{Real World Scenarios}
To collect our real-world dataset, we made a prototype on a real car. We used an IDS U3-36L0XC camera and an Ouster OS1-128 LiDAR (Rev 7). Regarding the HD headlight, we opt for the Digital Micromirror Device technology, which offer the greatest resolution ($<$0.015°) with a horizontal FOV of 14° and vertical FOV of 7°.  
All the hardware was mounted on the roof of the car. This novel setup allowed us to test LED under another projector-camera configuration. Ground truth annotation was made using LiDAR data, aggregated and densified following DOC-Depth method \cite{de2025doc} and thanks to Exwayz  software \cite{exwayz_3dm}. 
Due to the high resolution of the headlight we were able to collect data with smaller cell size than in simulation, giving more insight into pattern resolution impact. 
The performance of our model on this dataset demonstrate LED capabilities on complex real-world scenarios. We 
illustrate additional qualitative results in \cref{fig:sup_quali_real}.
A video of DepthFormer (LED 0.25°) qualitative results on our real-world dataset is available in the supplementary material.

\begin{figure}[t]
    \centering
     \includegraphics[width=0.9\linewidth]{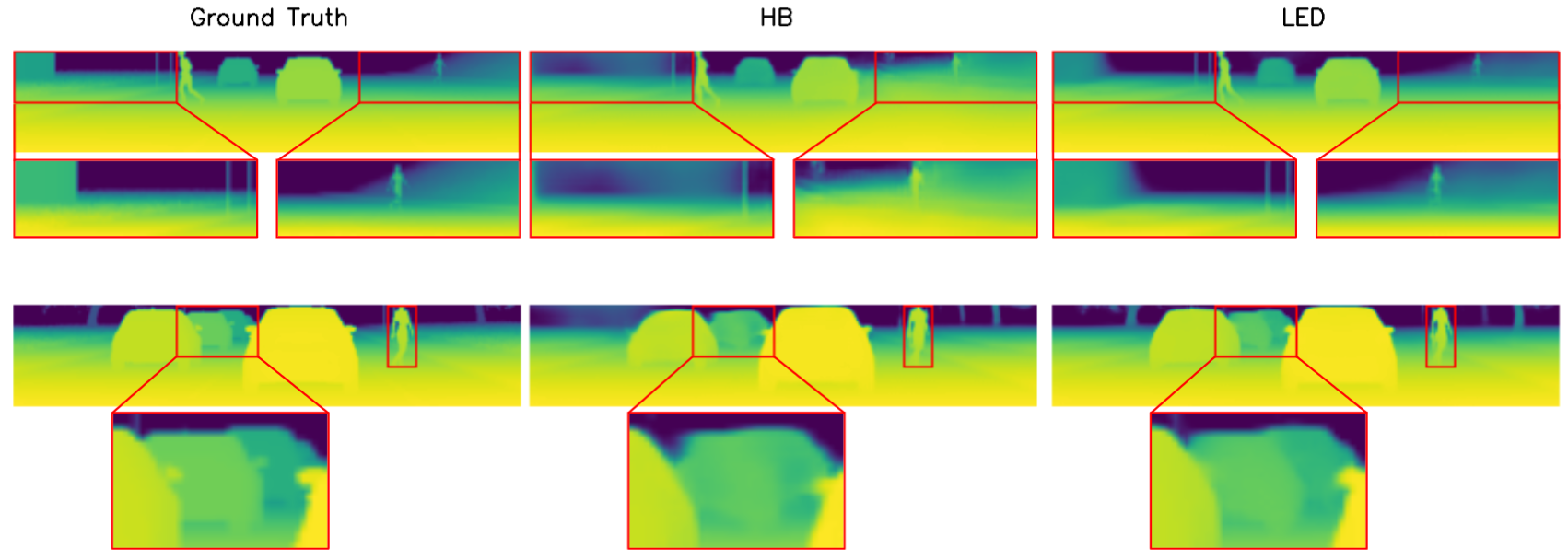}
     \caption{\textbf{Zoomed in qualitative results on NSDD} showcasing LED improved depth prediction compared to high-beam (HB).}
     \label{fig:sup_quali_synth}
\end{figure}
\section{Qualitative Results on Synthetic Dataset}
We illustrate in \cref{fig:sup_quali_synth} the improvements of depth estimation achieved by our method by comparing results with and without pattern. 
They are obtained by taking the encoder-decoder, trained on high beam (HB), and with our HD pattern (LED). 
Red boxes emphasize on enhanced regions. Some are zoomed in for improved visibility. 
LED results (right) exhibit higher precision, leading to more accurate object boundaries and shapes compared to HB (middle). Far away obstacles are better defined and less blurry (first row), vehicles and pedestrians are sharper (second rows).
\vspace{-5pt}

%\newpage
%\bibliography{supplementary}
%\end{document}

%% file: main.bbl
\begin{thebibliography}{52}
\providecommand{\natexlab}[1]{#1}
\providecommand{\url}[1]{\texttt{#1}}
\expandafter\ifx\csname urlstyle\endcsname\relax
  \providecommand{\doi}[1]{doi: #1}\else
  \providecommand{\doi}{doi: \begingroup \urlstyle{rm}\Url}\fi

\bibitem[Agarwal and Arora(2023)]{agarwal2023attention}
Ashutosh Agarwal and Chetan Arora.
\newblock Attention attention everywhere: Monocular depth prediction with skip attention.
\newblock In \emph{WACV}, 2023.

\bibitem[Baek and Heide(2021)]{baek_polka_2021}
Seung-Hwan Baek and Felix Heide.
\newblock Polka {Lines}: {Learning} {Structured} {Illumination} and {Reconstruction} for {Active} {Stereo}.
\newblock In \emph{CVPR}, 2021.

\bibitem[Baker et~al.(2020)Baker, Lu, Erlikhman, and Kellman]{baker2020local}
Nicholas Baker, Hongjing Lu, Gennady Erlikhman, and Philip~J Kellman.
\newblock Local features and global shape information in object classification by deep convolutional neural networks.
\newblock \emph{VR}, 2020.

\bibitem[Bhat et~al.(2021)Bhat, Alhashim, and Wonka]{bhat2021adabins}
Shariq~Farooq Bhat, Ibraheem Alhashim, and Peter Wonka.
\newblock Adabins: Depth estimation using adaptive bins.
\newblock In \emph{CVPR}, 2021.

\bibitem[Bian et~al.(2019)Bian, Li, Wang, Zhan, Shen, Cheng, and Reid]{bian2019unsupervised}
Jiawang Bian, Zhichao Li, Naiyan Wang, Huangying Zhan, Chunhua Shen, Ming-Ming Cheng, and Ian Reid.
\newblock Unsupervised scale-consistent depth and ego-motion learning from monocular video.
\newblock In \emph{NeurIPS}, 2019.

\bibitem[Bijelic et~al.(2018{\natexlab{a}})Bijelic, Gruber, and Ritter]{bijelic2018benchmark}
Mario Bijelic, Tobias Gruber, and Werner Ritter.
\newblock A benchmark for lidar sensors in fog: Is detection breaking down?
\newblock In \emph{IV}, 2018{\natexlab{a}}.

\bibitem[Bijelic et~al.(2018{\natexlab{b}})Bijelic, Gruber, and Ritter]{bijelic2018benchmarking}
Mario Bijelic, Tobias Gruber, and Werner Ritter.
\newblock Benchmarking image sensors under adverse weather conditions for autonomous driving.
\newblock In \emph{IV}, 2018{\natexlab{b}}.

\bibitem[Bochkovskii et~al.(2024)Bochkovskii, Delaunoy, Germain, Santos, Zhou, Richter, and Koltun]{depth_pro}
Aleksei Bochkovskii, Ama\"{e}l Delaunoy, Hugo Germain, Marcel Santos, Yichao Zhou, Stephan~R. Richter, and Vladlen Koltun.
\newblock Depth pro: Sharp monocular metric depth in less than a second.
\newblock \emph{arXiv}, 2024.

\bibitem[Brahmanage and Leung(2019)]{brahmanage_outdoor_2019}
Gayan Brahmanage and Henry Leung.
\newblock Outdoor {RGB}-{D} {Mapping} {Using} {Intel}-{RealSense}.
\newblock In \emph{{IEEE} Sensors}, 2019.

\bibitem[Caesar et~al.(2020)Caesar, Bankiti, Lang, Vora, Liong, Xu, Krishnan, Pan, Baldan, and Beijbom]{nuscenes}
Holger Caesar, Varun Bankiti, Alex~H. Lang, Sourabh Vora, Venice~Erin Liong, Qiang Xu, Anush Krishnan, Yu~Pan, Giancarlo Baldan, and Oscar Beijbom.
\newblock nuscenes: A multimodal dataset for autonomous driving.
\newblock In \emph{CVPR}, 2020.

\bibitem[Cheng et~al.(2021)Cheng, Misra, Schwing, Kirillov, and Girdhar]{cheng2021mask2former}
Bowen Cheng, Ishan Misra, Alexander~G. Schwing, Alexander Kirillov, and Rohit Girdhar.
\newblock Masked-attention mask transformer for universal image segmentation.
\newblock \emph{arXiv}, 2021.

\bibitem[De~Charette et~al.(2012)De~Charette, Tamburo, Barnum, Rowe, Kanade, and Narasimhan]{de_charette_fast_2012}
Raoul De~Charette, Robert Tamburo, Peter~C. Barnum, Anthony Rowe, Takeo Kanade, and Srinivasa~G. Narasimhan.
\newblock Fast reactive control for illumination through rain and snow.
\newblock In \emph{ICCP}, 2012.

\bibitem[de~Moreau et~al.(2025)de~Moreau, Mathias, Hassan, Yasser, Andrei, Hafid, Fabien, et~al.]{de2025doc}
Simon de~Moreau, Corsia Mathias, Bouchiba Hassan, Almehio Yasser, Bursuc Andrei, El-Idrissi Hafid, Moutarde Fabien, et~al.
\newblock Doc-depth: A novel approach for dense depth ground truth generation.
\newblock In \emph{IEEE IV}, 2025.

\bibitem[Exwayz(2024)]{exwayz_3dm}
Exwayz.
\newblock Exwayz 3d mapping: Create dense, accurate and georeferenced 3d point cloud using lidar slam.
\newblock \url{https://www.exwayz.fr/}, 2024.

\bibitem[Fanello et~al.(2016)Fanello, Rhemann, Tankovich, Kowdle, Escolano, Kim, and Izadi]{fanello_hyperdepth_2016}
Sean~Ryan Fanello, Christoph Rhemann, Vladimir Tankovich, Adarsh Kowdle, Sergio~Orts Escolano, David Kim, and Shahram Izadi.
\newblock {HyperDepth}: {Learning} {Depth} from {Structured} {Light} without {Matching}.
\newblock In \emph{CVPR}, 2016.

\bibitem[Gasperini et~al.(2023)Gasperini, Morbitzer, Jung, Navab, and Tombari]{gasperini_morbitzer2023md4all}
Stefano Gasperini, Nils Morbitzer, HyunJun Jung, Nassir Navab, and Federico Tombari.
\newblock Robust monocular depth estimation under challenging conditions.
\newblock In \emph{ICCV}, 2023.

\bibitem[Geiger et~al.(2012)Geiger, Lenz, and Urtasun]{Geiger2012CVPR}
Andreas Geiger, Philip Lenz, and Raquel Urtasun.
\newblock Are we ready for autonomous driving? the kitti vision benchmark suite.
\newblock In \emph{CVPR}, 2012.

\bibitem[Godard et~al.(2017)Godard, Mac~Aodha, and Brostow]{godard2017unsupervised}
Cl{\'e}ment Godard, Oisin Mac~Aodha, and Gabriel~J Brostow.
\newblock Unsupervised monocular depth estimation with left-right consistency.
\newblock In \emph{CVPR}, 2017.

\bibitem[Godard et~al.(2019)Godard, Mac~Aodha, Firman, and Brostow]{godard2019digging}
Cl{\'e}ment Godard, Oisin Mac~Aodha, Michael Firman, and Gabriel~J Brostow.
\newblock Digging into self-supervised monocular depth estimation.
\newblock In \emph{ICCV}, 2019.

\bibitem[Guizilini et~al.(2020)Guizilini, Ambrus, Pillai, Raventos, and Gaidon]{guizilini20203d}
Vitor Guizilini, Rares Ambrus, Sudeep Pillai, Allan Raventos, and Adrien Gaidon.
\newblock 3d packing for self-supervised monocular depth estimation.
\newblock In \emph{CVPR}, 2020.

\bibitem[Gupta et~al.(2013)Gupta, Yin, and Nayar]{gupta_structured_2013}
Mohit Gupta, Qi~Yin, and Shree~K. Nayar.
\newblock Structured {Light} in {Sunlight}.
\newblock In \emph{ICCV}, 2013.

\bibitem[Hu et~al.(2019)Hu, Ozay, Zhang, and Okatani]{HigherResMap}
Junjie Hu, Mete Ozay, Yan Zhang, and Takayuki Okatani.
\newblock Revisiting single image depth estimation: Toward higher resolution maps with accurate object boundaries.
\newblock In \emph{WACV}, 2019.

\bibitem[Ke et~al.(2024)Ke, Obukhov, Huang, Metzger, Daudt, and Schindler]{marigold}
Bingxin Ke, Anton Obukhov, Shengyu Huang, Nando Metzger, Rodrigo~Caye Daudt, and Konrad Schindler.
\newblock Repurposing diffusion-based image generators for monocular depth estimation.
\newblock In \emph{CVPR}, 2024.

\bibitem[Lee et~al.(2019)Lee, Han, Ko, and Suh]{lee2019big}
Jin~Han Lee, Myung-Kyu Han, Dong~Wook Ko, and Il~Hong Suh.
\newblock From big to small: Multi-scale local planar guidance for monocular depth estimation.
\newblock \emph{arXiv preprint arXiv:1907.10326}, 2019.

\bibitem[Li et~al.(2022)Li, Monno, and Okutomi]{li_deep_2022}
Chunyu Li, Yusuke Monno, and Masatoshi Okutomi.
\newblock Deep {Hyperspectral}-{Depth} {Reconstruction} {Using} {Single} {Color}-{Dot} {Projection}.
\newblock In \emph{CVPR}, 2022.

\bibitem[Li(2022)]{lidepthtoolbox2022}
Zhenyu Li.
\newblock Monocular depth estimation toolbox.
\newblock \url{https://github.com/zhyever/Monocular-Depth-Estimation-Toolbox}, 2022.

\bibitem[Li et~al.(2023)Li, Chen, Liu, and Jiang]{li2023depthformer}
Zhenyu Li, Zehui Chen, Xianming Liu, and Junjun Jiang.
\newblock Depthformer: Exploiting long-range correlation and local information for accurate monocular depth estimation.
\newblock \emph{MIR}, 2023.

\bibitem[Liu et~al.(2021)Liu, Song, Wang, Liu, and Zhang]{liu2021self}
Lina Liu, Xibin Song, Mengmeng Wang, Yong Liu, and Liangjun Zhang.
\newblock Self-supervised monocular depth estimation for all day images using domain separation.
\newblock In \emph{ICCV}, 2021.

\bibitem[Loshchilov and Hutter(2018)]{loshchilov2017decoupled}
Ilya Loshchilov and Frank Hutter.
\newblock Decoupled weight decay regularization.
\newblock In \emph{ICLR}, 2018.

\bibitem[Maddern et~al.(2017)Maddern, Pascoe, Linegar, and Newman]{RobotCarDatasetIJRR}
Will Maddern, Geoff Pascoe, Chris Linegar, and Paul Newman.
\newblock {1 Year, 1000km: The Oxford RobotCar Dataset}.
\newblock \emph{IJRR}, 2017.

\bibitem[Mertz et~al.(2012)Mertz, Koppal, Sia, and Narasimhan]{mertz_low-power_2012}
Christoph Mertz, Sanjeev~J. Koppal, Solomon Sia, and Srinivasa Narasimhan.
\newblock A low-power structured light sensor for outdoor scene reconstruction and dominant material identification.
\newblock In \emph{CVPR Workshops}, 2012.

\bibitem[Nathan~Silberman and Fergus(2012)]{Silberman:ECCV12}
Pushmeet~Kohli Nathan~Silberman, Derek~Hoiem and Rob Fergus.
\newblock Indoor segmentation and support inference from rgbd images.
\newblock In \emph{ECCV}, 2012.

\bibitem[NVIDIA(2024)]{drivesim}
NVIDIA.
\newblock Nvidia drive sim.
\newblock \url{https://www.nvidia.com/en-us/self-driving-cars/simulation/}, 2024.
\newblock Accessed: 2024-01-18.

\bibitem[Paszke et~al.(2019)Paszke, Gross, Massa, Lerer, Bradbury, Chanan, Killeen, Lin, Gimelshein, Antiga, Desmaison, Kopf, Yang, DeVito, Raison, Tejani, Chilamkurthy, Steiner, Fang, Bai, and Chintala]{Pytorch_NEURIPS2019_9015}
Adam Paszke, Sam Gross, Francisco Massa, Adam Lerer, James Bradbury, Gregory Chanan, Trevor Killeen, Zeming Lin, Natalia Gimelshein, Luca Antiga, Alban Desmaison, Andreas Kopf, Edward Yang, Zachary DeVito, Martin Raison, Alykhan Tejani, Sasank Chilamkurthy, Benoit Steiner, Lu~Fang, Junjie Bai, and Soumith Chintala.
\newblock Pytorch: An imperative style, high-performance deep learning library.
\newblock In \emph{NeurIPS}, 2019.

\bibitem[Poggi et~al.(2018)Poggi, Aleotti, Tosi, and Mattoccia]{poggi2018towards}
Matteo Poggi, Filippo Aleotti, Fabio Tosi, and Stefano Mattoccia.
\newblock Towards real-time unsupervised monocular depth estimation on cpu.
\newblock In \emph{IROS}, 2018.

\bibitem[Riegler et~al.(2019)Riegler, Liao, Donne, Koltun, and Geiger]{riegler_connecting_2019}
Gernot Riegler, Yiyi Liao, Simon Donne, Vladlen Koltun, and Andreas Geiger.
\newblock Connecting the {Dots}: {Learning} {Representations} for {Active} {Monocular} {Depth} {Estimation}.
\newblock In \emph{CVPR}, 2019.

\bibitem[Ronneberger et~al.(2015)Ronneberger, Fischer, and Brox]{U-Net}
Olaf Ronneberger, Philipp Fischer, and Thomas Brox.
\newblock U-net: Convolutional networks for biomedical image segmentation.
\newblock In \emph{MICCAI}, 2015.

\bibitem[Shu et~al.(2020)Shu, Yu, Duan, and Yang]{shu2020feature}
Chang Shu, Kun Yu, Zhixiang Duan, and Kuiyuan Yang.
\newblock Feature-metric loss for self-supervised learning of depth and egomotion.
\newblock In \emph{ECCV}, 2020.

\bibitem[Silberman and Fergus(2011)]{silberman_indoor_2011}
Nathan Silberman and Rob Fergus.
\newblock Indoor scene segmentation using a structured light sensor.
\newblock In \emph{ICCV Workshops}, 2011.

\bibitem[Spencer et~al.(2020)Spencer, Bowden, and Hadfield]{spencer2020defeat}
Jaime Spencer, Richard Bowden, and Simon Hadfield.
\newblock Defeat-net: General monocular depth via simultaneous unsupervised representation learning.
\newblock In \emph{CVPR}, 2020.

\bibitem[Tamburo et~al.(2014)Tamburo, Nurvitadhi, Chugh, Chen, Rowe, Kanade, and Narasimhan]{fleet_programmable_2014}
Robert Tamburo, Eriko Nurvitadhi, Abhishek Chugh, Mei Chen, Anthony Rowe, Takeo Kanade, and Srinivasa~G. Narasimhan.
\newblock Programmable {Automotive} {Headlights}.
\newblock In \emph{ECCV}, 2014.

\bibitem[Vankadari et~al.(2020)Vankadari, Garg, Majumder, Kumar, and Behera]{vankadari2020unsupervised}
Madhu Vankadari, Sourav Garg, Anima Majumder, Swagat Kumar, and Ardhendu Behera.
\newblock Unsupervised monocular depth estimation for night-time images using adversarial domain feature adaptation.
\newblock In \emph{ECCV}, 2020.

\bibitem[Waldner and Bertram(2021)]{waldner_optimal_2021}
Mirko Waldner and Torsten Bertram.
\newblock Optimal {Real}-time {Digitization} of {Matrix}-{Headlights}.
\newblock In \emph{AIM}, 2021.

\bibitem[Waldner et~al.(2019)Waldner, Kramer, and Bertram]{waldner_hardware---loop-simulation_2019}
Mirko Waldner, Maximilian Kramer, and Torsten Bertram.
\newblock Hardware-in-the-{Loop}-{Simulation} of the light distribution of automotive {Matrix}-{LED}-{Headlights}.
\newblock In \emph{AIM}, 2019.

\bibitem[Waldner et~al.(2020)Waldner, Kramer, and Bertram]{waldner_digitization_2020}
Mirko Waldner, Maximilian Kramer, and Torsten Bertram.
\newblock Digitization of {Matrix}-{Headlights} {That} {Move} as in the {Real} {Test} {Drive}.
\newblock In \emph{AIM}, 2020.

\bibitem[Waldner et~al.(2022)Waldner, Müller, and Bertram]{waldner_energy-efficient_2022}
Mirko Waldner, Nathalie Müller, and Torsten Bertram.
\newblock Energy-{Efficient} {Illumination} by {Matrix} {Headlamps} for {Nighttime} {Automated} {Object} {Detection}.
\newblock \emph{IJECER}, 2022.

\bibitem[Wang et~al.(2021)Wang, Zhang, Yan, Li, Xu, Li, and Yang]{wang2021regularizing}
Kun Wang, Zhenyu Zhang, Zhiqiang Yan, Xiang Li, Baobei Xu, Jun Li, and Jian Yang.
\newblock Regularizing nighttime weirdness: Efficient self-supervised monocular depth estimation in the dark.
\newblock In \emph{CVPR}, 2021.

\bibitem[Yang et~al.(2024{\natexlab{a}})Yang, Kang, Huang, Xu, Feng, and Zhao]{depth_anything_v1}
Lihe Yang, Bingyi Kang, Zilong Huang, Xiaogang Xu, Jiashi Feng, and Hengshuang Zhao.
\newblock Depth anything: Unleashing the power of large-scale unlabeled data.
\newblock In \emph{CVPR}, 2024{\natexlab{a}}.

\bibitem[Yang et~al.(2024{\natexlab{b}})Yang, Kang, Huang, Zhao, Xu, Feng, and Zhao]{depth_anything_v2}
Lihe Yang, Bingyi Kang, Zilong Huang, Zhen Zhao, Xiaogang Xu, Jiashi Feng, and Hengshuang Zhao.
\newblock Depth anything v2.
\newblock \emph{arXiv:2406.09414}, 2024{\natexlab{b}}.

\bibitem[Yin and Shi(2018)]{yin2018geonet}
Zhichao Yin and Jianping Shi.
\newblock Geonet: Unsupervised learning of dense depth, optical flow and camera pose.
\newblock In \emph{CVPR}, 2018.

\bibitem[Zheng et~al.(2023)Zheng, Zhong, Li, Gao, Zheng, Jin, Wang, Zhao, Zhou, Zhang, et~al.]{zheng2023steps}
Yupeng Zheng, Chengliang Zhong, Pengfei Li, Huan-ang Gao, Yuhang Zheng, Bu~Jin, Ling Wang, Hao Zhao, Guyue Zhou, Qichao Zhang, et~al.
\newblock Steps: Joint self-supervised nighttime image enhancement and depth estimation.
\newblock In \emph{ICRA}, 2023.

\bibitem[Zhou et~al.(2017)Zhou, Brown, Snavely, and Lowe]{zhou2017unsupervised}
Tinghui Zhou, Matthew Brown, Noah Snavely, and David~G Lowe.
\newblock Unsupervised learning of depth and ego-motion from video.
\newblock In \emph{CVPR}, 2017.

\end{thebibliography}
